\documentclass[letterpaper, 10 pt, conference]{ieeeconf}  

\IEEEoverridecommandlockouts                              

\usepackage{graphics,subfigure} 
\usepackage{epsfig} 
\usepackage{amsmath} 
\usepackage{amssymb}  
\usepackage[usenames,dvipsnames]{color}
\usepackage{multirow}
\usepackage{hyperref}       

\graphicspath {{figure/}}

\newcommand{\argmax}{\operatornamewithlimits{argmax}}
\newcommand{\argmin}{\operatornamewithlimits{argmin}}

\newtheorem{thm}{Theorem}

\title{\LARGE \bf
Probabilistic Framework for Constrained Manipulations and \\Task and Motion Planning under Uncertainty
}

\author{Jung-Su Ha and Danny Driess and Marc Toussaint
\thanks{All authors are with the Machine Learning \& Robotics Lab, University Stuttgart and with the Max Planck Institute for Intelligent Systems, Stuttgart, Germany
        {\tt\small jung-su.ha@ipvs.uni-stuttgart.de}}%
}

\begin{document}

\maketitle
\thispagestyle{empty}
\pagestyle{empty}

\begin{abstract}
Logic-Geometric Programming (LGP) is a powerful motion and manipulation planning framework, which represents hierarchical structure using logic rules that describe discrete aspects of problems, e.g., touch, grasp, hit, or push, and solves the resulting smooth trajectory optimization. The expressive power of logic allows LGP for handling complex, large-scale sequential manipulation and tool-use planning problems. In this paper, we extend the LGP formulation to stochastic domains. Based on the control-inference duality, we interpret LGP in a stochastic domain as fitting a mixture of Gaussians to the posterior path distribution, where each logic profile defines a single Gaussian path distribution. The proposed framework enables a robot to prioritize various interaction modes and to acquire interesting behaviors such as contact exploitation for uncertainty reduction, eventually providing a composite control scheme that is reactive to disturbance. The supplementray video can be found at \href{https://youtu.be/CEaJdVlSZyo}{https://youtu.be/CEaJdVlSZyo}.
\end{abstract}

\section{Introduction}
Manipulation planning problems often involve two major difficulties, namely high-dimensionality and discontinuous contact dynamics, which prohibit widely-used motion planning algorithms such as sampling-based planning~\cite{lavalle1998rapidly,karaman2011sampling} or trajectory optimization~\cite{mayne1966second,todorov2005generalized,toussaint2017tutorial} from being directly applicable.
To handle such difficulties, hybrid approaches have been proposed, where additional variables that explicitly represent discrete aspects of problems are incorporated into optimization and jointly optimized~\cite{mordatch2012discovery,deits2014footstep,toussaint2018differentiable}. For example, contact invariant optimization~\cite{mordatch2012discovery} relaxes the discontinuity of contact dynamics and utilizes the additional continuous-valued variable to express contact activity that enforces the trajectory to be consistent with physics. In~\cite{deits2014footstep}, the additional integer variables describe hybrid contact activities and the resulting Mixed-Integer Programming is solved with optimization algorithms involving branch-and-bound. Logic-Geometric Programming (LGP)~\cite{toussaint2018differentiable} adopts logic rules to describe discrete aspects of problems on a higher level than typical contact activities, e.g., touch, hit, push, or more general tool-use. A sequence of these logic states (called a \textit{skeleton}) directly implies contact activities over time, which imposes equality/inequality constraints for smooth trajectory optimization. The expressive power of logic enables LGP to enumerate valuable local optima of the planning problem by searching over the logic space.

\begin{figure}[t]
	\centering
	\subfigure{
		\includegraphics[width=.99\columnwidth, viewport = 50 200 2400 790, clip]{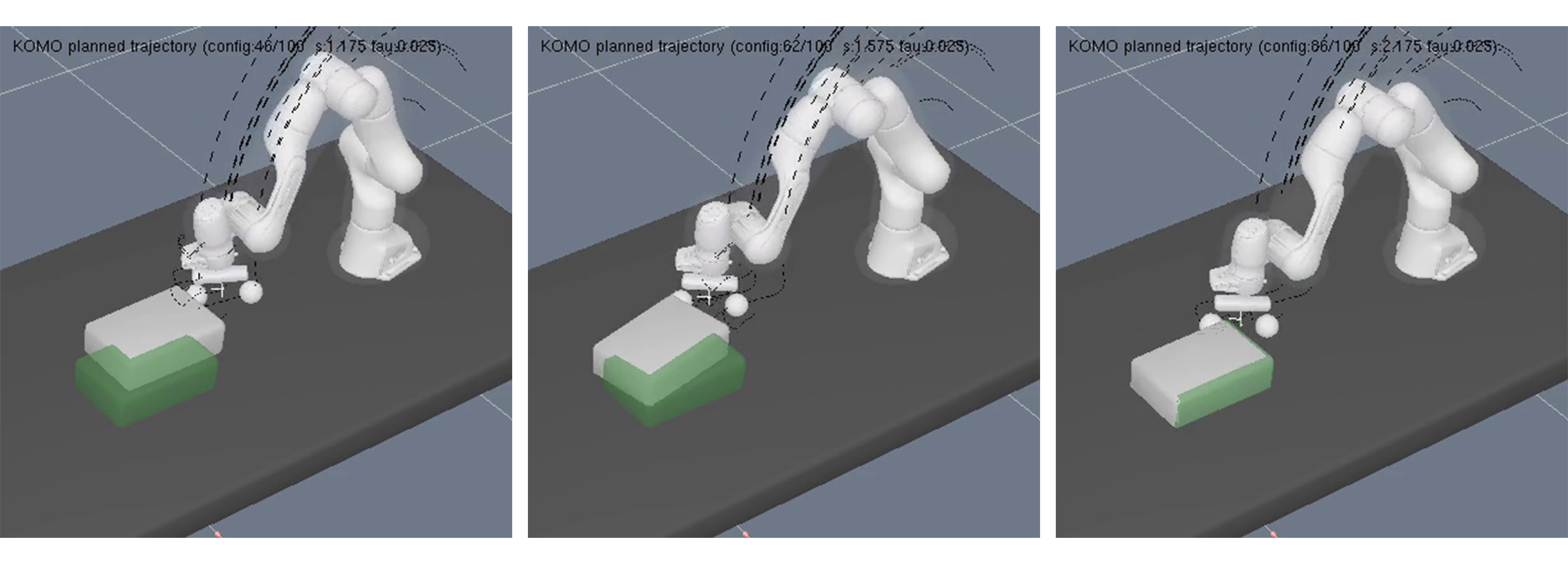}}
	\subfigure{
		\includegraphics[width=.99\columnwidth, viewport = 50 200 2400 790, clip]{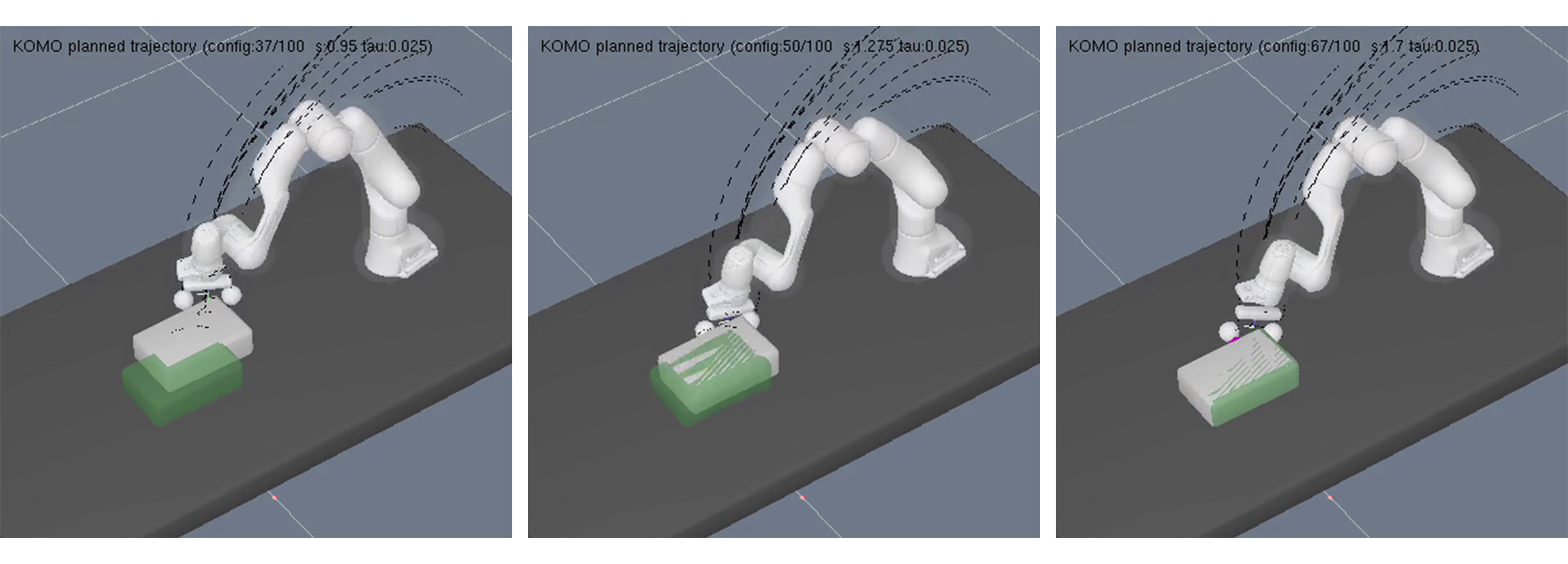}}
	\vspace{-0.7cm}
	\caption{Two strategies for box pushing. A robot can use one finger (upper), or two fingers (lower). Obviously, the latter strategy is more robust to disturbance, thereby incurring smaller feedback control cost. }\label{fig:push}
	\vspace{-0.6cm}
\end{figure}
In this work, we present a probabilistic framework of such hybrid trajectory optimization by extending LGP to stochastic domains, where the dynamics is described stochastically and the cost function is the expectation over all possible trajectories. The corresponding problems can be formulated as stochastic optimal control (SOC), which gives rise to some important and interesting features. First, when comparing various local optima (plans) with different skeletons, the robustness of the plans should be taken into account. For example, consider a planning problem in Fig. \ref{fig:push} whose objective is to push some object on a table towards target area using a single finger or two fingers. Both plans might incur similar costs in a deterministic sense, but the single-finger push strategy is much less favorable in reality because it is more vulnerable to disturbances and uncertainties. Second, we can observe contact exploitation behaviors. To cope with actuator disturbance, a robot might want to fix some parts of its body, e.g. elbow, on the desk to reduce the uncertainty of end-effector's position. Lastly, depending on the deviation from the plan, a robot might decide whether or not to switch to another plan or even to stay in-between them. The original LGP formulation is only deterministic so, even though it can find various feasible plans, a robot cannot help but choose one plan to execute based on the deterministic path cost. In contrast, the probabilistic framework in this work allows a robot for prioritizing different plans by taking robustness as well as deterministic path cost into account and for acquiring interesting contact exploitation behaviors. Furthermore, a composite reactive controller constructed from various plans enables a robot to adaptively choose which plan(s) to follow.

The technical aspect of this work is based on the duality between control and inference~\cite{todorov2008general,rawlik2012stochastic,kappen2012optimal}. Under this duality, motion planning is equivalent to inference of posterior path distribution. As in trajectory optimization, the gradient and Hessian can accelerate the inference procedure, which relates to the Laplace approximation~\cite{toussaint2009robot}. Given the fact that the prescribed logics provide smoothness of sub-problems, we interpret LGP in stochastic domains as fitting a mixture of Gaussians to the posterior path distribution, where each skeleton defines a single Gaussian path distribution.


\section{Background}
\subsection{Stochastic optimal control as KL-minimization}
Consider the configuration space $\mathcal{X}=\mathbb{R}^{d_x}$ of an $d_x$-dimensional robot and an initial configuration $x_0\in\mathcal{X}$ and a velocity $\dot{x}_0$ are given.
Let $\mathbf{z}=(x, \dot{x})\in\mathbb{R}^{2d_x}$ be a state vector, $\mathbf{u}\in\mathbb{R}^{d_u}$ be a control vector which represents torques or desired accelerations of actuated joints ($d_u\leq d_x$), and $\mathbf{w}$ be a $d_u$-dimensional Wiener process that is injected into the robot's actuators.
Then the robot dynamics can be written as the following stochastic differential equation (SDE), which is affine in the control input and the disturbance:
\begin{equation}
d\mathbf{z}(t) =  \mathbf{f}(\mathbf{z}(t))dt+ G(\mathbf{z}(t))(\mathbf{u}(t)dt + \sigma d\mathbf{w}(t)), \label{eq:conti_dyn}
\end{equation}
where $\mathbf{f}:\mathbb{R}^{2d_x}\rightarrow \mathbb{R}^{2d_x}$ is the passive dynamics and $G:\mathbb{R}^{2d_x}\rightarrow \mathbb{R}^{2d_x\times d_u}$ is the control transition matrix function.
With an instantaneous state cost rate $V: \mathbb{R}^{2d_x} \rightarrow \mathbb{R}^+$, an SOC problem is formulated as follow:
\begin{equation}
J = \mathbb{E}_{p_\mathbf{u}}\!\left[V_T(\mathbf{z}(T))+\!\!\int^{T}_0\!\!\! V(\mathbf{z}(t))+\frac{1}{2\sigma^2}\mathbf{u}(t)^T\mathbf{u}(t)dt\right]\!\!,\!\! \label{eq:conti_cost}
\end{equation}
where $p_\mathbf{u}$ is the probability measure induced by the controlled trajectories in \eqref{eq:conti_dyn}, with $\mathbf{z}(0) = \mathbf{z}_0 = (x_0, \dot{x}_0)$. The objective of an SOC problem is then to find a control policy $\mathbf{u}(t) = \pi^*(\mathbf{z}(t),t)$ that minimizes the cost functional~\eqref{eq:conti_cost}.

The above types of SOC problems, which are defined with control/disturbance-affine dynamics and quadratic control cost, are called linearly solvable optimal control problems and have interesting properties~\cite{todorov2009efficient,kappen2012optimal}. In particular, they can be transformed into Kullback-Leibler (KL) divergence minimization~\cite{kappen2016adaptive,ha2018adaptive} using the following theorem.
\begin{thm}[Girsanov's Theorem~\cite{gardiner1985handbook}]\textit{
	Suppose $p_\mathbf{0}$ is the probability measures induced by the uncontrolled trajectories from \eqref{eq:conti_dyn} with $\mathbf{z}(0) = (x_0, \dot{x}_0)$ and $\mathbf{u}(t)=0~\forall t\in[0, T]$.
	Then, the Radon-Nikodym derivative of $p_\mathbf{u}$ with respect to $p_\mathbf{0}$ is given by
	\begin{align}
	\frac{dp_\mathbf{u}}{dp_\mathbf{0}} = \exp\left(\frac{1}{2\sigma^2}\!\int^{T}_0\!\!\!||\mathbf{u}(t)||^2dt +\frac{1}{\sigma}\!\int^{T}_0\!\!\mathbf{u}(t)^Td\mathbf{w}(t)\right)\!\!,\! \label{eq:Radon}
	\end{align}
	where $\mathbf{w}(t)$ is a Wiener process for $p_\mathbf{u}$.}
\end{thm}
With Girsanov's theorem, the objective function~\eqref{eq:conti_cost} can be rewritten in terms of KL divergence:
\begin{align}
J&=\mathbb{E}_{p_\mathbf{u}}\left[V_T(\mathbf{z}(T))+\int_{0}^{T}V(\mathbf{z}(t))+\frac{1}{2\sigma^2}||\mathbf{u}(t)||^2dt\right]\nonumber\\
&=\mathbb{E}_{p_\mathbf{u}}\left[V_T(\mathbf{z}(T))+\int_{0}^{T}V(\mathbf{z}(t))dt+\log\frac{dp_\mathbf{u}(\mathbf{z}_{[0,T]})}{dp_\mathbf{0}(\mathbf{z}_{[0,T]})}\right]\nonumber\\
&=\mathbb{E}_{p_\mathbf{u}}\left[\log\frac{dp_\mathbf{u}(\mathbf{z}_{[0,T]})}{dp_\mathbf{0}(\mathbf{z}_{[0,T]})\exp(-V(\mathbf{z}_{[0,T]}))/\xi}-\log\xi\right] \nonumber\\
&=D_\text{KL}\left(p_\mathbf{u}(\mathbf{z}_{[0,T]})||p^*(\mathbf{z}_{[0,T]})\right)-\log\xi,
\end{align}
where $\mathbf{z}_{[0,T]}\equiv\{\mathbf{z}(t);~\forall t\in[0, T]\}$ is a state trajectory, $V(\mathbf{z}_{[0,T]})\equiv V_T(\mathbf{z}(T))+\int^{T}_{0}V(\mathbf{z}(t))dt$ is a trajectory state cost and $\xi \equiv \int\exp(-V(\mathbf{z}_{[0,T]}))dp_\mathbf{0}(\mathbf{z}_{[0,T]})$ is a normalization constant.\footnote{Note that the second term in the exponent of \eqref{eq:Radon} disappears when taking expectation w.r.t. $p_\mathbf{u}$, i.e. $\mathbb{E}_{p_\mathbf{u}}[\int_{0}^{T}\mathbf{u}(t)^Td\mathbf{w}(t)]=0$.}
Because $\xi$ is a constant, $p^*(\mathbf{z}_{[0,T]})$ can be interpreted as the optimally-controlled trajectory distribution that minimizes the cost functional \eqref{eq:conti_cost}:
\begin{align}
dp^*(\mathbf{z}_{[0,T]}) &= \frac{\exp(-V(\mathbf{z}_{[0,T]}))dp_\mathbf{0}(\mathbf{z}_{[0,T]})}{\int\exp(-V(\mathbf{z}_{[0,T]}))dp_\mathbf{0}(\mathbf{z}_{[0,T]})}, \label{eq:opt_dist}
\end{align}
and the corresponding optimal cost is given by 
\begin{align}
J^* = -\log\int\exp(-V(\mathbf{z}_{[0,T]}))dp_\mathbf{0}(\mathbf{z}_{[0,T]}). \label{eq:opt_cost}
\end{align}

Once the optimal trajectory distribution is obtained, the optimal control can be recovered by enforcing the controlled dynamics to mimic the optimal trajectory distribution, e.g., via moment matching. By applying Girsanov's theorem to \eqref{eq:opt_dist}, the optimal trajectory distribution can be expressed as:
\begin{align}
dp^*(\mathbf{z}_{[0,T]}) \propto dp_\mathbf{u}(\mathbf{z}_{[0,T]})\exp\left(-V_\mathbf{u}(\mathbf{z}_{[0,T]})\right), \label{eq:opt_dist2}
\end{align}
where $V_\mathbf{u}(\mathbf{z}_{[0,T]}) = V(\mathbf{z}_{[0,T]})+\frac{1}{2\sigma^2}\int^{T}_0||\mathbf{u}(t)||^2dt +\int^{T}_0\mathbf{u}(t)^Td\mathbf{w}(t).$ The SOC framework that utilizes the importance sampling scheme to approximate this distribution is called path integral control~\cite{ha2016topology,kappen2016adaptive}. It samples a set of trajectories $\mathbf{z}_{[0,T]}^{l}\sim p_\mathbf{u}(\cdot)$, assigns their importance weights as $\tilde{w}^{l}\propto\exp(-V_\mathbf{u}(\mathbf{z}_{[0,T]}^{l}))$, and computes the optimal control by matching the first (and second) moments of $p_\mathbf{u}$ to $p^*$.

\subsection{Laplace approximation of path distributions}
Instead of relying on sampling schemes for approximating $p^*$, this work builds on the efficient local optimization methods by investigating a close connection between second order trajectory optimization algorithms~\cite{mayne1966second,todorov2005generalized,toussaint2017tutorial}\footnote{Our method especially builds upon the framework of \textit{k}-order Motion Optimization (KOMO)~\cite{toussaint2017tutorial} which has the same efficiency as the others while being able to address more general problems.} and the Laplace approximation. 
The Laplace approximation fits a normal distribution to the first two derivatives of the log target density function at the mode. 
Let $\mathbf{x}=x_{1:N}=(x_1,x_2,...,x_N)\in\mathbb{R}^{N\times d_x}$ be a path representation of $\mathbf{z}_{[0,T]}$, which is a path of $N$ time steps in the configuration space $\mathcal{X}$.
In this path representation, the velocity and acceleration (and control inputs) of the joints can be computed from two and three consecutive configurations, respectively, using the finite difference approximation.\footnote{The path representation significantly reduces the size of optimization problems, which leads to better numerical stability~\cite{erez2012trajectory,zucker2013chomp,toussaint2017tutorial}.}
Slightly abusing the notation, the uncontrolled path distribution and the state trajectory cost are then expressed as functions of $\mathbf{x}$:
\begin{align}
p_\mathbf{0}(\mathbf{x}) &\propto \exp\left(-\sum^{N}_{n=1}f_\mathbf{0}(x_{n-2:n})\right), \\
\exp\left(-V(\mathbf{x})\right) &= \exp\left(-\sum^{N}_{n=1}f_V(x_{n-1:n})\right),
\end{align}
for an appropriately given prefix $x_{-1:0}$.
Then, the problem of finding the mode $\mathbf{x}^*$ of $p^*(\mathbf{x}) \propto p_\mathbf{0}(\mathbf{x})\exp\left(-V(\mathbf{x})\right)$ is an unconstrained nonlinear program (NLP):
\begin{align}
\min_{x_{1:N}} \sum^N_{n=1}f_\mathbf{0}(x_{n-2:n})+f_V(x_{n-1:n}), \label{eq:LA_uncon}
\end{align}
which can be solved using the Newton-Raphson algorithm, i.e., $\mathbf{x}^{i+1} = \mathbf{x}^{i} - {\nabla^2f(\mathbf{x}^{i})}^{-1}\nabla f(\mathbf{x}^{i})$ where $f(\mathbf{x}) = \sum^N_{n=1}f_\mathbf{0}(x_{n-2:n})+f_V(x_{n-1:n})$.
With a solution, $\mathbf{x}^*$, and a Hessian at the solution, $\nabla^2f(\mathbf{x}^*)$, the resulting Laplace approximation is given by:
\begin{align}
p^*(\mathbf{x}) \approx \mathcal{N}(\mathbf{x}|\mathbf{x}^*, {\nabla^2f(\mathbf{x}^*)}^{-1}). \label{eq:LA_result}
\end{align} 
The optimal cost is also approximated similarly from~\eqref{eq:opt_cost}:
\begin{align}
J^* \approx f(\mathbf{x}^*)-\frac{1}{2}\log\frac{|{\nabla^2f_\mathbf{0}(\mathbf{x}^*)}|}{|{\nabla^2f(\mathbf{x}^*)}|}. \label{eq:LA_result_cost}
\end{align}
Note that the covariance of the approximate distribution has full rank for fully-actuated robots ($d_u = d_x$), but not for underacted robots ($d_u<d_x$). In such cases, the NLP should be formulated with equality constraints that restrict the uncontrollable subspace to be consistent with the dynamics \eqref{eq:conti_dyn}. Details will be addressed in Section~\ref{sec:PLGP_mixture}.

\section{Estimating the Path Distribution for Constrained Trajectory Optimization}
\subsection{Logic geometric programming in stochastic domains}
We now consider more general manipulation planning problems where the configuration space $\mathcal{X}=\mathbb{R}^{d_x}\times SE(3)^m$ involves $m$ rigid objects as well as a $d_x$-dimensional robot.
The dynamics in~\eqref{eq:conti_dyn} becomes complicated in this case, because it should express interactions between the robot and the objects. The objects are controllable only when they are in contact with the robot, thus the dynamics~\eqref{eq:conti_dyn} is discontinuous around the contact activities. Local optimization methods are no longer effective since they cannot utilize the well-defined gradient and Hessian along the directions of contact switching. 
Logic Geometric Programming (LGP) addresses this difficulty by augmenting the formulation with additional logic decision variables, $(s_{1:K}, a_{1:K})$, which describe discrete aspects of dynamics in a higher level, e.g., touch, hit, push, or more general tool-use. A mode $s_k$ imposes a set of equality/inequality constraints for the prescribed contact activities to the optimization while a switch $a_k$ represents transitions between the modes.
We can formulate an LGP problem in stochastic domains as follows:
\begin{equation}
\min_{\substack{\mathbf{u}_{[0,T]},\\ a_{1:K}, s_{1:K}}} E_{p_\mathbf{u}}\left[V_T(\mathbf{z}(T))+\int^{T}_0 V(\mathbf{z}(t))+\frac{1}{2\sigma^2}||\mathbf{u}(t)||^2dt\right] \nonumber
\end{equation}
\begin{align}
\text{s.t.}~&\forall t\in[0,T]:~h_\text{path}(\mathbf{z}(t),s_{k(t)})=0,~g_\text{path}(\mathbf{z}(t),s_{k(t)})\leq 0 \nonumber\\
&d\mathbf{z}(t) =  \mathbf{f}_{s_{k(t)}}(\mathbf{z}(t))dt+ G_{s_{k(t)}}(\mathbf{z}(t))(\mathbf{u}(t)dt + \sigma d\mathbf{w}(t))\nonumber\\
&\forall^K_{k=1}:~h_\text{switch}(\mathbf{z}(t),a_{k(t)})=0,~g_\text{switch}(\mathbf{z}(t),a_{k(t)})\leq 0,\nonumber\\
&~~~~~~~~~s_k\in\text{succ}(s_{k-1},a_k). \label{eq:const_SOC}
\end{align}
Here, the SDE is conditioned on $s$ so that it can be defined only in the remaining subspace which is not constrained by the path constraints, $(h,g)_\text{path}$.
For example, when a mode specifies manipulation of a particular object, the SDE represents the robot's dynamics constrained for that specified interaction and the dynamics of the manipulated object is defined by path constraints.
Because the contact activities are prescribed by the \textit{skeleton} $a_{1:K}$ and the smooth switch constraints $(h,g)_\text{switch}$, the dynamics in~\eqref{eq:const_SOC} is now smooth w.r.t. the state and the control inputs, thereby making the corresponding SOC, $\mathcal{P}(a_{1:K})$, to be smooth.

LGP problems are often addressed with a two-level hierarchical approach~\cite{toussaint2017multi,toussaint2018differentiable}, where a higher-level module proposes a skeleton $a_{1:K}$ using, e.g., tree search and a lower-level NLP solver returns a solution of $\mathcal{P}(a_{1:K})$. LGP in stochastic domains~\eqref{eq:const_SOC} has two distinctive features: While $\mathcal{P}(a_{1:K})$ is evaluated for a single optimal trajectory in the deterministic case, the path distribution should instead be considered, which results in the additional stochastic cost term (Sec. \ref{sec:PLGP_mixture}); also, considering various modes allows for constructing the composite reactive control law (Sec. \ref{sec:composition}).

\subsection{Probabilistic LGP as fitting a mixture of Gaussians} \label{sec:PLGP_mixture}
Let $\{\mathbf{a}_i={a_{1:K}}^{(i)}; i=1,...,N_a\}$ be a set of candidate skeletons for an LGP problem~\eqref{eq:const_SOC}.
We now attempt to approximate the optimal path distribution as a mixture distribution, where each skeleton defines one mixture component:
\begin{align}
p^*(\mathbf{x})&\approx{\sum}_{i=1}^{N_a} p^*(\mathbf{a}_i) p^*(\mathbf{x}|\mathbf{a}_i). \label{eq:opt_dist_mixture}
\end{align}
The mixture component $p^*(\mathbf{x}|\mathbf{a}_i)$ corresponds to the SOC problem $\mathcal{P}(\mathbf{a}_i)$, of which support is defined by the constraints of the original problem, i.e.,:
\begin{align}
&p^*(\mathbf{x}|\mathbf{a}_i)\propto p_\mathbf{0}(\mathbf{x}|\mathbf{s}_i)\exp(-V(\mathbf{x})) \nonumber\\
&\text{s.t.}~h_\text{path}(\mathbf{x},\mathbf{s})=0,~g_\text{path}(\mathbf{x},\mathbf{s})\leq 0, \nonumber\\
&~~~~h_\text{switch}(\mathbf{x},\mathbf{a})=0,~g_\text{switch}(\mathbf{x},\mathbf{a})\leq 0,\nonumber\\
&~~~~\forall^K_{k=1}:~s_k\in\text{succ}(s_{k-1},a_k).
\end{align}
Given that $\mathcal{P}(a_{1:K})$ is a \textit{smooth} SOC problem, we can use the Laplace approximation to represent each mixture component $p^*(\mathbf{x}|\mathbf{a}_i)$ as a Gaussian distribution.
Apart from the unconstrained NLP in \eqref{eq:LA_uncon}, however, $\mathcal{P}(a_{1:K})$ yields a more complicated trajectory optimization problem; e.g., the dynamics of moving or manipulated objects are defined by equality constraints and the resting objects are just imposing inequality constraints for collision avoidance.
Such problems should be formulated as a constrained NLP:
\begin{align}
& \min_{x_{1:N}} \sum^N_{n=1}f_\mathbf{0}(x_{n-2:n})+f_V(x_{n-1:n}) \nonumber\\
\text{s.t.}~&\forall^N_{n=1}: h(x_{n-1:n})=0,~g(x_{n-1:n})\leq 0, \label{eq:LA_con}
\end{align}
which can be addressed by any constrained optimization methods, such as Augmented Lagrangian Gauss-Newton.

Suppose we have found $\mathbf{x}^*_i$, an NLP solution for the $i^\text{th}$ skeleton. We then approximate the $i^\text{th}$ mixture density as:
\begin{align}
p^*(\mathbf{x}|\mathbf{a}_i) \approx \mathcal{N}(\mathbf{x}|\mathbf{x}^*_i, \Sigma^*_i). \label{eq:LA_con_result}
\end{align}
Because of the equality/inequality constraints imposed in \eqref{eq:LA_con}, this distribution is degenerate; i.e., deviations from $\mathbf{x}^*_i$ can only lie in the kernel of the equality and active inequality constraint Jacobians, thereby making the above distribution only span a lower-dimensional subspace.
Let a column of matrix $W$ denote an orthonormal basis of the nullspace of $J=\begin{bmatrix}\nabla h(\mathbf{x}^*)\\\nabla \text{diag}(\lambda)g(\mathbf{x}^*)\end{bmatrix}$, where $\lambda$ is the dual variables on the inequality constraints. Then, $\Sigma^*_i$ is given by the inverse of the projected second derivatives of $\log p^*(\mathbf{x})$ at $\mathbf{x}^*$:
\begin{align}
\Sigma^*_i=W\left(W^T\nabla^2f(\mathbf{x}^*_i)W\right)^{-1}W^T. \label{eq:Sigstar}
\end{align}


To complete the mixture approximation, we also need to compute the mixture weights, $p^*(\mathbf{a}_i)$. Because a skeleton imposes different constraints to the corresponding NLP~\eqref{eq:LA_con}, making each mixture component span different subspaces, we can assume that the modes are widely separated, which enables the mixture weights to be computed independently \cite[Chapter 12]{gelman2013bayesian}:
\begin{align}
p^*(\mathbf{a}_i) &\propto p_\mathbf{0}(\mathbf{x}^*_i|\mathbf{s}_i)\exp(-V(\mathbf{x}^*_i))\{(2\pi)^{d}|\Sigma^*_i|_+\}^{1/2} \nonumber\\
 &\propto \exp\left(-f(\mathbf{x}^*_i)\right)\left(|\Sigma^*_i|_+/|\Sigma_i|_+\right)^{1/2}, \label{eq:mixture_weight}
\end{align}
where $\Sigma_i=W\left(W^T\nabla^2f_\mathbf{0}(\mathbf{x}^*_i)W\right)^{-1}W^T$ is the covariance of the (degenerate) uncontrolled path distribution $p_\mathbf{0}(\mathbf{x}|\mathbf{a}_i)$, $d=\text{rank}(\Sigma^*_i)$, and $|\cdot|_+$ denotes a pseudo-determinant.
Note that the mixture weights~\eqref{eq:mixture_weight} are determined by two factors. The first term is a cost of the optimal deterministic path which are exponentially penalized. The second term can be interpreted as a stochastic cost that penalizes an entropy ratio between the optimal and uncontrolled path distributions; 
robust plans have large margins for deviations from the reference so the controlled path distribution need not be shrunk via feedback control, while plans vulnerable to disturbance do not allow even a small deviation, requiring a high-gain feedback controller to make the controlled path distribution relatively narrow. In other words, this term represents the expected \textit{feedback} control cost from the optimal controller. 
See the equivalence between the optimal value of the unimodal approximation~\eqref{eq:LA_result_cost} and the log of the mixture weights~\eqref{eq:mixture_weight};
the mass of each mixture component is assigned according to the total (deterministic + stochastic) cost incurred by that plan.
Finally, the multimodal approximation of the optimal cost is given by:
\begin{align}
J^* \approx -\log\sum_{i=1}^{N_a}\frac{1}{N_a}\exp\left(-f(\mathbf{x}^*_i)\right)\frac{|\Sigma^*_i|_+^{1/2}}{|\Sigma_i|_+^{1/2}}, \label{eq:MA_result_cost}
\end{align}
which, of course, becomes the cost in~\eqref{eq:LA_result_cost} when $N_a=1$.

\section{Reactive Controller for Mode Switching}
After planning, we need a control policy to execute the plan. Using the multi-modal path distribution, the control policy should be able to not only stabilize a particular reference trajectory, but also decide which reference trajectory to follow. This section is devoted to derive the stabilizing controllers within each mode (Sec. \ref{sec:KODP}) and to introduce two methods to synthesize those controllers (Sec. \ref{sec:composition}).

\subsection{$k$-order constrained dynamic programming}\label{sec:KODP}
Within each mode, to derive the controller for general constrained cases, consider the following recursive equation: 
\begin{align}
&J_n(x_{n-2:n-1})\nonumber\\
&=\min_{x_{n:N}} \sum^N_{l=n}f_l(x_{l-2:l})~\text{s.t.}~\forall^N_{l=n}: h_l=0,~g_l\leq 0 \label{eq:value}\\
&=\min_{x_{n}} \big[f_n(x_{n-2:n})+J_{n+1}(x_{n-1:n})\big]~\text{s.t.}~h_n=0,~g_n\leq 0, \nonumber
\end{align}
where $J_{N+1} \equiv 0$. Such procedures for computing the cost-to-go function $J_n$ is called $k$-order constrained dynamic programming (KODP)~\cite[Sec. 4]{toussaint2017tutorial}, which is an generalization of the Bellman optimality equation to the $k$-order (2nd-order in~\eqref{eq:value}) and constrained cases.

In particular, the linear feedback controller for a computed plan can be built directly from the gradient and Hessian of the optimization. Let $\delta x = x - x^*$ and consider the quadratic and linear approximations of $J$, $f$, $h$, and $g$:
\begin{align}
&J_n(x_{n-2:n-1}) \equiv \frac{1}{2}\delta x^TV_n\delta x + v_n^T\delta x + \bar{v} \nonumber\\
&f_n(x_{n-2:n-1}) \approx \frac{1}{2}\delta x^T\nabla^2f^*\delta x + (\nabla f^*)^T\delta x + f^* \nonumber\\
&h_n(x_{n-2:n-1}) \approx (\nabla h^*)^T\delta x,~g_n(x_{n-2:n-1}) \approx (\nabla g^*)^T\delta x. \nonumber
\end{align}
Note that all the gradients (and Hessian) of $f,~h,~g$ are already computed while solving the NLP~\eqref{eq:LA_con}.\footnote{We can also include additional penalties like $f \gets f + \rho||x-x^*||^*$ or modify the weights between cost terms to adjust the closed loop behaviors.}
The minimization in KODP~\eqref{eq:value} is then written as:
\begin{align}
\delta x_n^*=&\argmin_{x_n} f(\delta x_{n-2:n})+J_{n+1}(\delta x_{n-1:n})\nonumber\\
&\text{s.t.}~h_n(\delta x_{n-2:n})=0,~g_n(\delta x_{n-2:n})\leq 0,\label{eq:Bellman}
\end{align}
which has the form of a quadratic program (QP) with\footnote{We leave out the inequality constraints for the sake of notation but the activated inequalities should be treated as the equality constraints.}
\begin{align}
&f_n + J_{n+1} \equiv \frac{1}{2}\begin{bmatrix}\delta x_{n-2:n-1}\\ \delta x_n\end{bmatrix}^T\begin{bmatrix}
D_n & C_n \\ C_n^T & E_n\end{bmatrix}\begin{bmatrix}\delta x_{n-2:n-1}\\ \delta x_n\end{bmatrix} \nonumber\\
 &~~~~~~~~~~~~~~~+ \begin{bmatrix}d_n\\ e_n\end{bmatrix}\begin{bmatrix}\delta x_{n-2:n-1}\\ \delta x_n\end{bmatrix} + c_n,\nonumber\\
&\text{s.t.}~\begin{bmatrix}l_n\\ m_n\end{bmatrix}^T\begin{bmatrix}\delta x_{n-2:n-1}\\ \delta x_n\end{bmatrix} = 0. 
\end{align}
Given the cost-to-go function at the next time step $J_{n+1}$, the solution of the above QP can be represented linearly around $\delta x = 0$ (which corresponds to the solution of the original problem) using the sensitivity analysis of NLPs~\cite{levy1995sensitivity,amos2017optnet}:
\begin{align}
&\begin{bmatrix}E_n & m_n \\ m_n^T & 0\end{bmatrix}\begin{bmatrix}\delta x_n^*\\ \delta \lambda_n^*\end{bmatrix}= \begin{bmatrix}-C_n^T\delta x_{n-2:n-1}-e_n\\ -l_n^T\delta x_{n-2:n-1}\end{bmatrix}. \label{eq:KKT}
\end{align}
The above directly implies the linear feedback control law:
\begin{align}
\begin{bmatrix}\delta x_n^*\\ \delta \lambda_n^*\end{bmatrix} &= -\begin{bmatrix}E_n & m_n \\ m_n^T & 0\end{bmatrix}^{-1}(\begin{bmatrix}e_n\\0\end{bmatrix}+\begin{bmatrix}C_n^T\\l_n^T\end{bmatrix}\delta x_{n-2:n-1})\nonumber\\
&=u^{ff}_n + K_n\delta x_{n-2:n-1}. \label{eq:opt_con}
\end{align}

The cost-to-go functions along the whole time horizon can be derived from the Bellman equation $J_n = \min_{\delta x_n}\big[f_n + J_{n+1}\big]$ which results in the following backward matrix recursion:
\begin{align}
V_n =& D_n + \frac{1}{2}\begin{bmatrix}C_n &  l_n\end{bmatrix}\bar{H}_n\begin{bmatrix}C_n^T\\l_n^T\end{bmatrix}-\begin{bmatrix}C_n &  l_n\end{bmatrix}H_n\begin{bmatrix}C_n^T\\0\end{bmatrix}\nonumber\\
v_n =& d_n - \begin{bmatrix}C_n &  0\end{bmatrix}H_n\begin{bmatrix}e_n\\0\end{bmatrix}+\begin{bmatrix}C_n &  l_n\end{bmatrix}\big(\bar{H}_n-H_n\big)\begin{bmatrix}e_n\\0\end{bmatrix} \nonumber\\
\bar{v}_n =& c_n + \frac{1}{2}\begin{bmatrix}e_n^T & 0\end{bmatrix} \big(\bar{H}_n - 2H_n\big) \begin{bmatrix}e_n\\0\end{bmatrix}, \label{eq:KODP}
\end{align}
where $H_n \equiv \begin{bmatrix}E_n & m_n \\ m_n^T & 0\end{bmatrix}^{-1}$ and $\bar{H}_n \equiv H_n\begin{bmatrix}E_n & 0 \\ 0 & 0\end{bmatrix}H_n$. Note that, in the 1st-order unconstrained case, the above recursion~\eqref{eq:KODP} is equivalent to the Riccati equation of LQR.

\begin{figure}[t]
	\centering
	\subfigure[]{
		\includegraphics[width=.225\columnwidth, viewport = 10 10 340 280, clip]{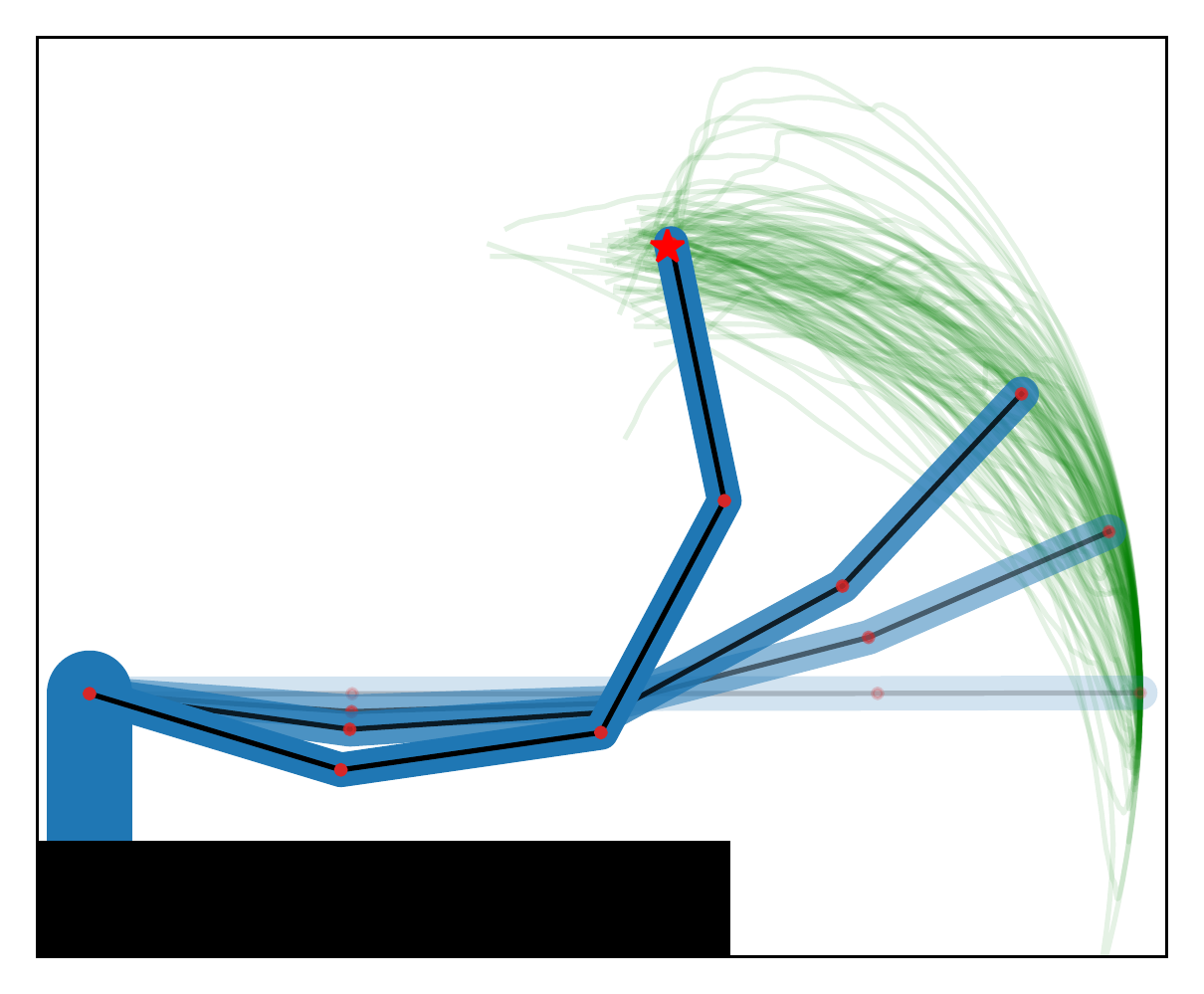}}	
	\subfigure[]{
		\includegraphics[width=.225\columnwidth, viewport = 10 10 340 280, clip]{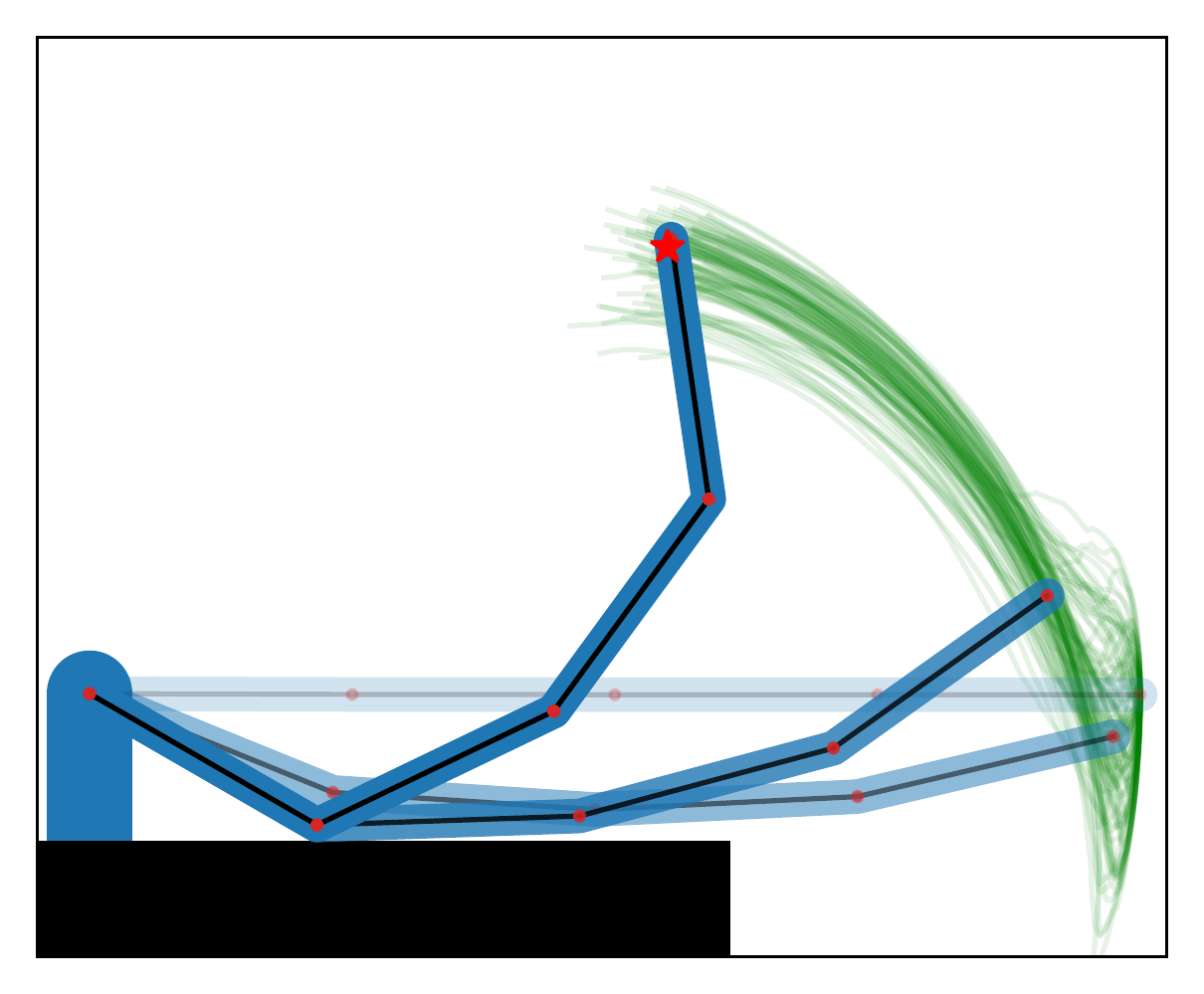}}
	\subfigure[]{
		\includegraphics[width=.225\columnwidth, viewport = 10 10 340 280, clip]{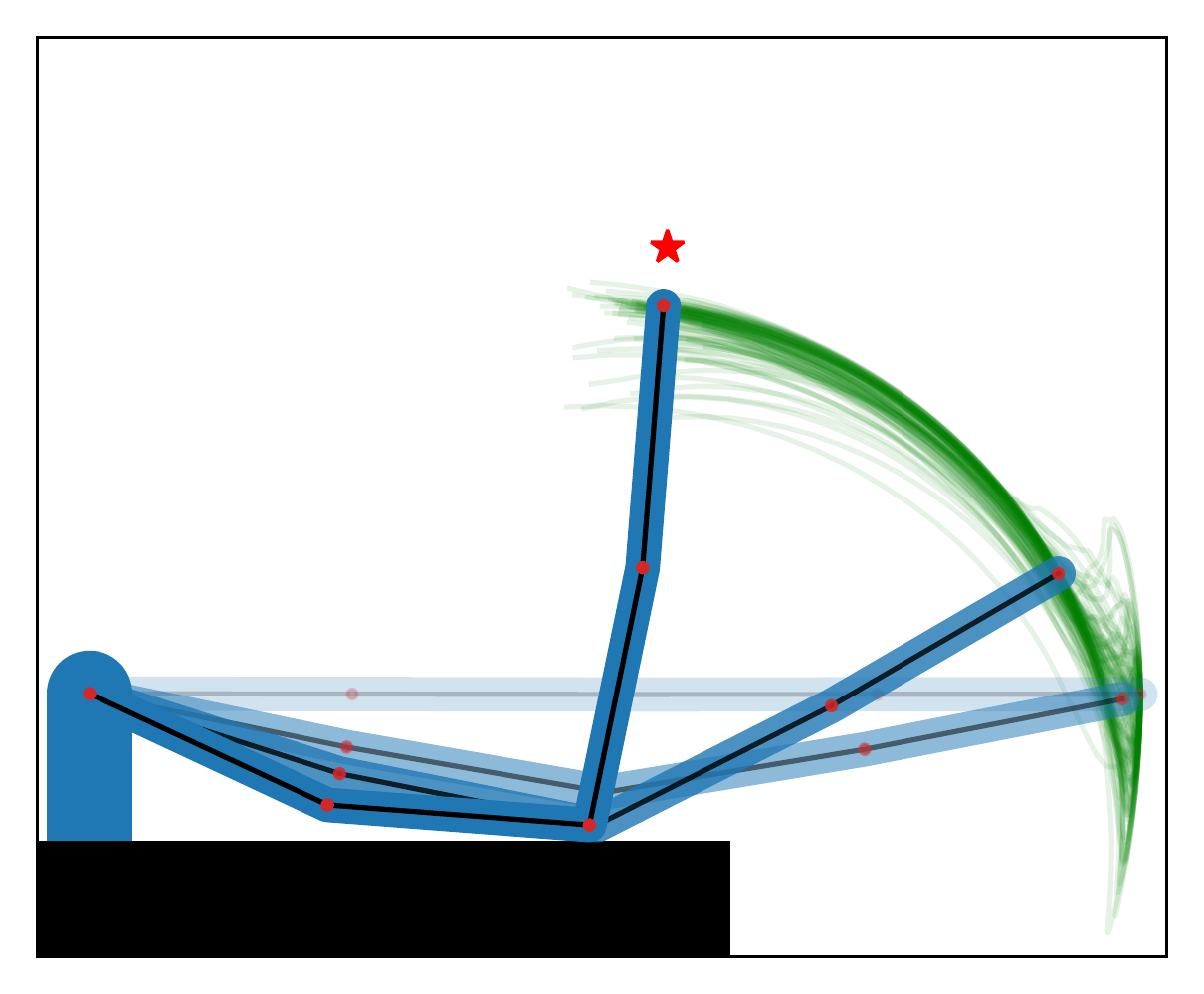}}
	\subfigure[]{
		\includegraphics[width=.225\columnwidth, viewport = 10 10 340 280, clip]{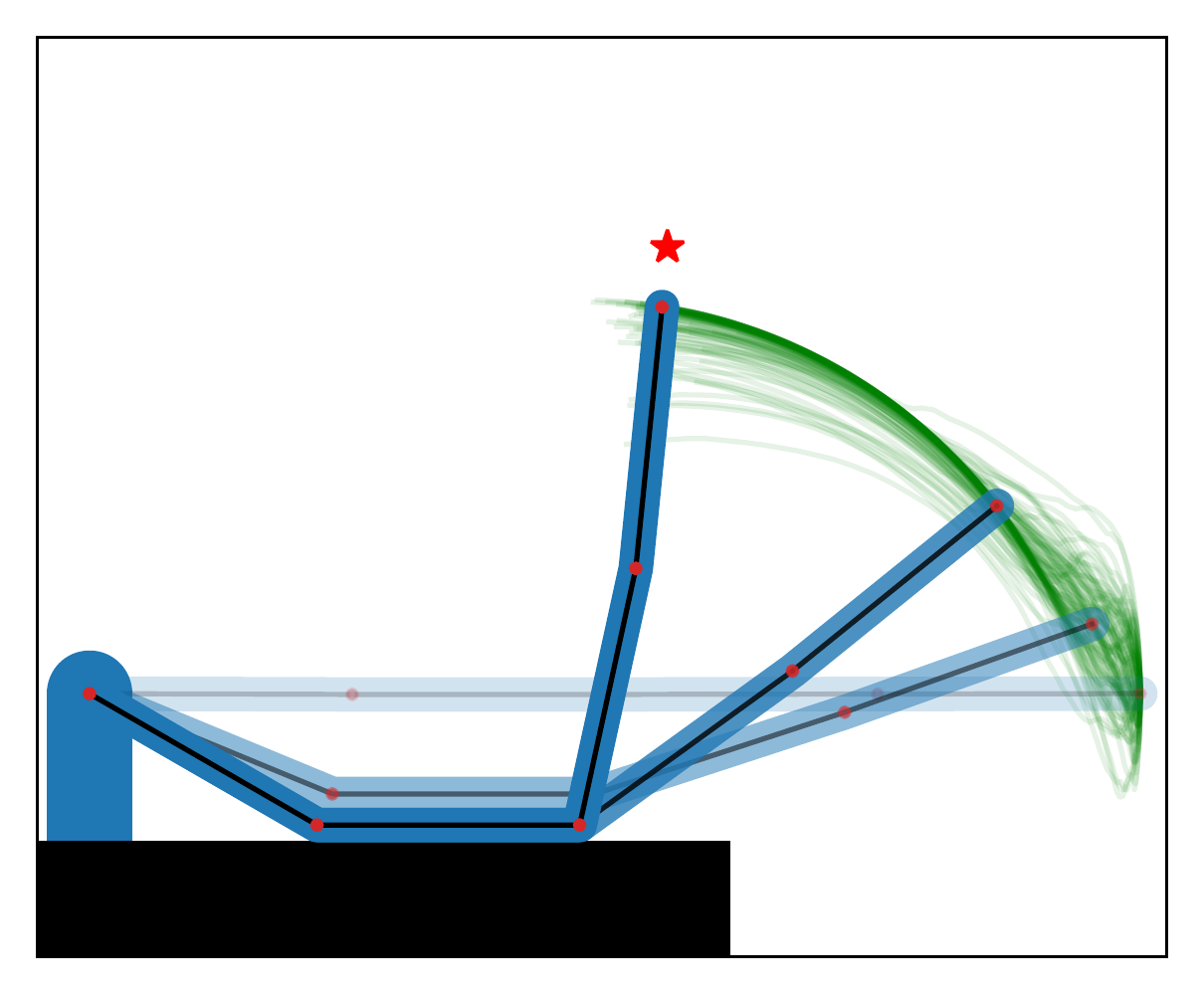}}\vspace*{-.3cm}
	\caption{Four different strategies for reaching task. The snapshots of the optimized path and sampled trajectories from the optimal path distributions.}\label{fig:reach}\vspace*{-.5cm}
\end{figure}
\subsection{Composite optimal control policy}\label{sec:composition}
During the execution, the mixture weights can be updated \textit{on the fly}. Let $J_n^{(i)}$ be the cost-to-go function w.r.t. an $i$th skeleton at a time step $l=n$ and $\tilde{\Sigma}$ be a covariance matrix that is only for the future trajectory $l=n,...,N$ which only takes submatrix of $\nabla^2f(\mathbf{x}^*_i)$ or $\nabla^2f_\mathbf{0}(\mathbf{x}^*_i)$.
Then, the mixture weight of an $i$th skeleton is given as:
\begin{align}
p^*_n(\mathbf{a}_i) \propto \exp\left(-J_n^{(i)}(x_{n-2:n-1})\right)\left(|\tilde{\Sigma}^*_i|_+/|\tilde{\Sigma}_i|_+\right)^{1/2} \label{eq:mixture_weight_onthefly}
\end{align}
where $x_{n-2:n-1}$ is the two past configurations that the robot observed. With these mixture weights, we introduce two different methods to construct the composite control policies from all skeletons, $u^{(i)}_n$ in \eqref{eq:opt_con}.
\begin{itemize}
	\item \textbf{Blending}: As suggested in \cite{todorov2009compositionality,muico2011composite}, the control input can be computed as a linear combination, i.e.,
	\begin{align}
	u^*_n = \sum\mathop{}_ip_n^*(\mathbf{a}_i)u^{(i)}_n, \label{eq:blending}
	\end{align}
	which minimizes forward KL divergence $D_\text{KL}(p^*||p_\mathbf{u})$.
	
	\item \textbf{Switching}: The blending method can cause undesirable smoothing effects in practice because it mixes behaviors for different contact activities. An alternative is to simply take the best expected policy as: 
	\begin{align}
	u^*_n = u^{(i^*)}_n,~i^* = \argmax_i p^*(\mathbf{a}_i). \label{eq:switching}
	\end{align}
\end{itemize}
We briefly show the different resulting behaviors of two methods in the following section.

The overall framework for planning and control can be summarized as follows: (i) In the offline planning phase, trajectories w.r.t. the different candidate skeletons proposed by a logic-level planner are optimized by solving~\eqref{eq:LA_con}, and for each skeleton, the cost-to-go function as well as the linear feedback policy are computed from KODP~\eqref{eq:KODP} and \eqref{eq:opt_con}. (ii) In the online execution phase, the mixture weights of the skeletons are assigned as~\eqref{eq:mixture_weight_onthefly} and, based on those, the control input is computed via \eqref{eq:blending} or \eqref{eq:switching}.
By considering various candidate plans and appropriately building composite policies, a robot can not only choose a more efficient and robust plan but also flexibly react to disturbance (e.g., stay in the current plan or switch to another).

\section{Demonstration}
We demonstrate our approach on three manipulation planning problems, reaching a target, pushing an object, and touching a banana. For clearer visualization and more results, we refer the readers to the supplementary video at \href{https://youtu.be/CEaJdVlSZyo}{https://youtu.be/CEaJdVlSZyo}.
\begin{figure}[t]
	\centering
	\subfigure[Single-finger push]{
		\includegraphics[width=.4\columnwidth]{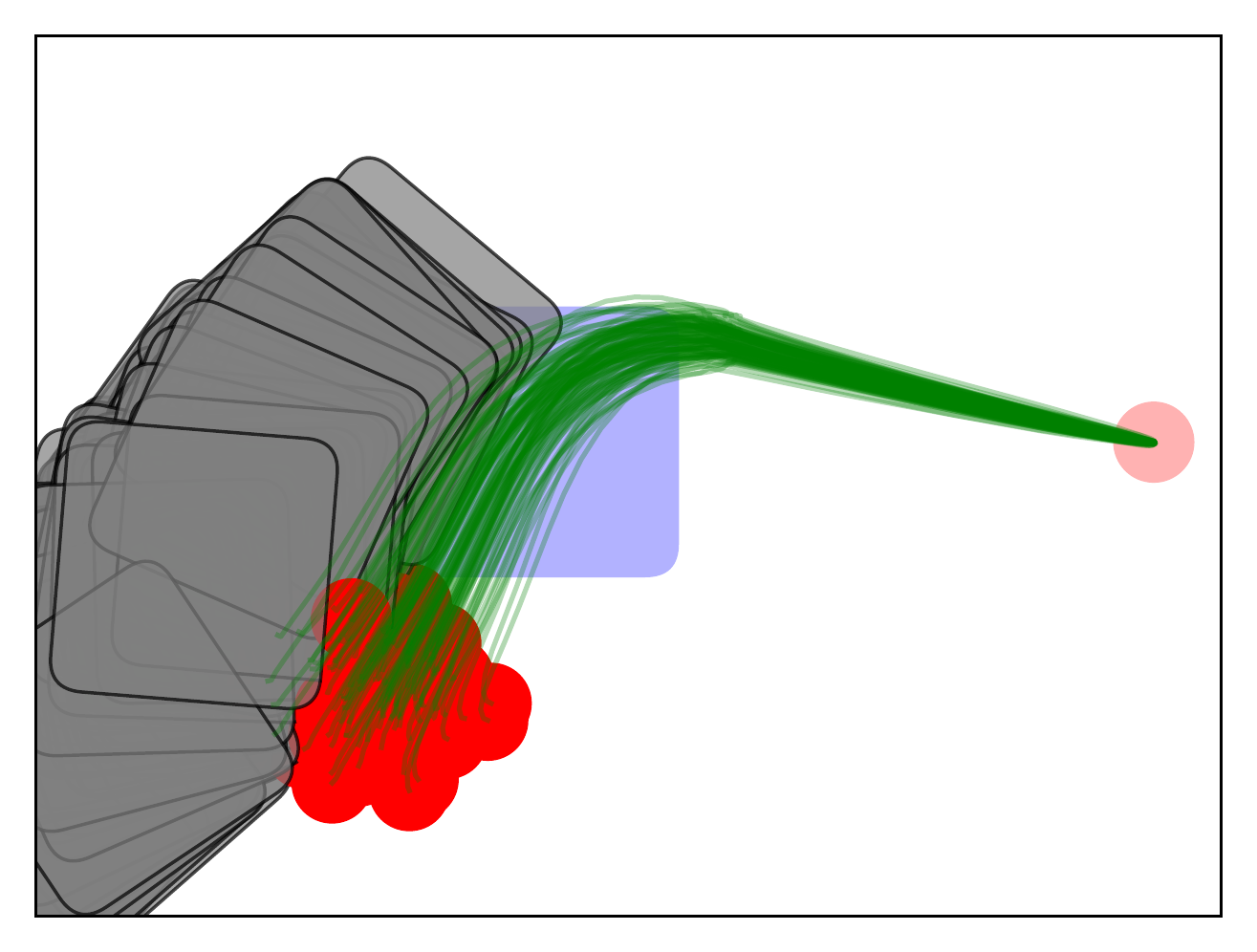}}
	\subfigure[Two-finger push]{
		\includegraphics[width=.4\columnwidth]{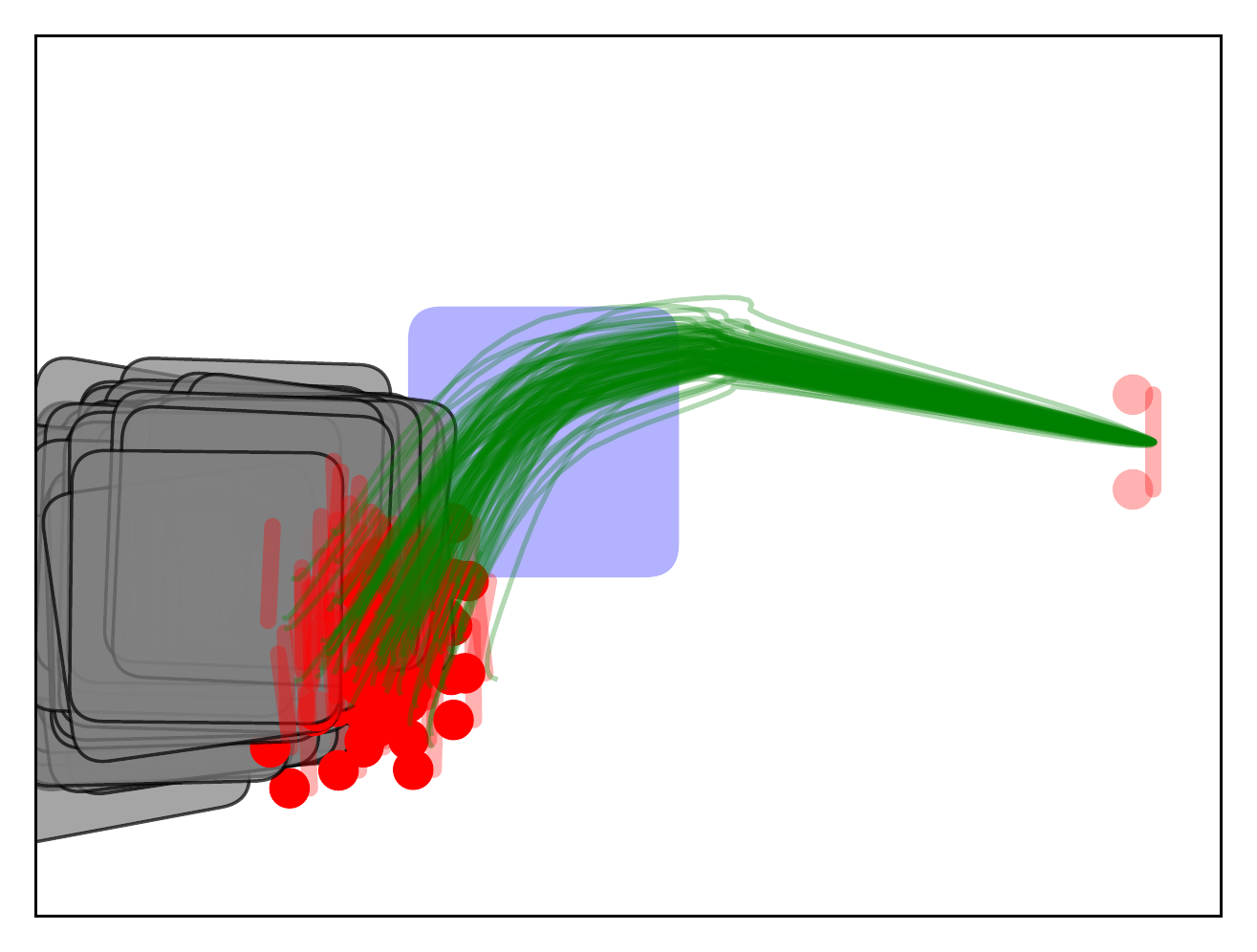}}\vspace*{-.3cm}
	\caption{Sample trajectories from $\mathcal{N}(\mathbf{x}^*,\Sigma)$, i.e., without feedback, for pushing. For the same level of disturbances, object's final configurations in the single-finger case are much more diverged from the target. The RMS errors are $0.3727$ and $0.0952$, respectively.}\label{fig:push_comp}\vspace*{-.3cm}
\end{figure}
\begin{figure}[t]
	\centering
	\subfigure[]{
		\includegraphics[width=.35\columnwidth, viewport = 10 10 340 280, clip]{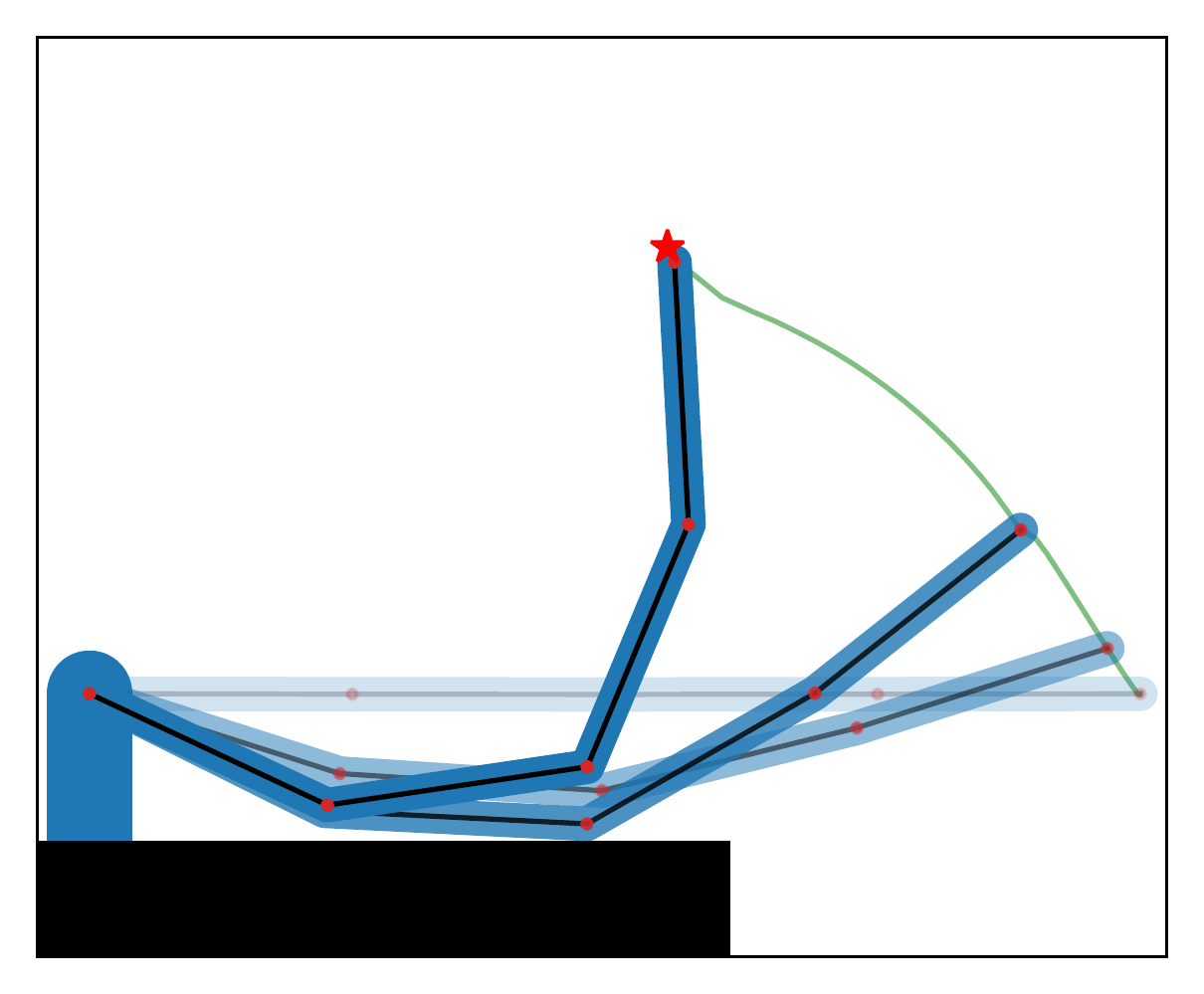}}	
	\subfigure[]{
		\includegraphics[width=.35\columnwidth, viewport = 10 10 340 280, clip]{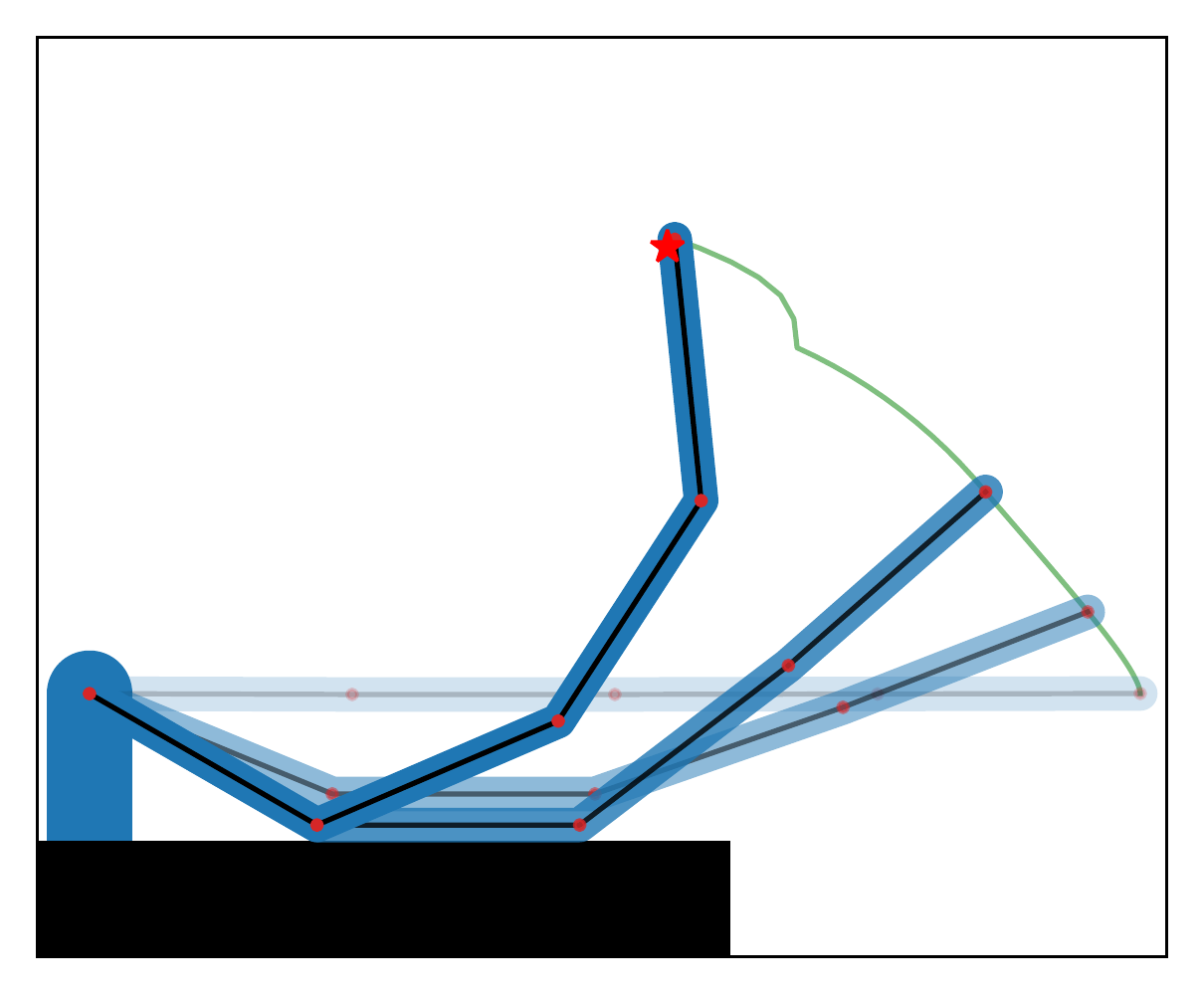}}	\\
	\subfigure[]{
		\includegraphics[width=.45\columnwidth, viewport = 5 10 400 315, clip]{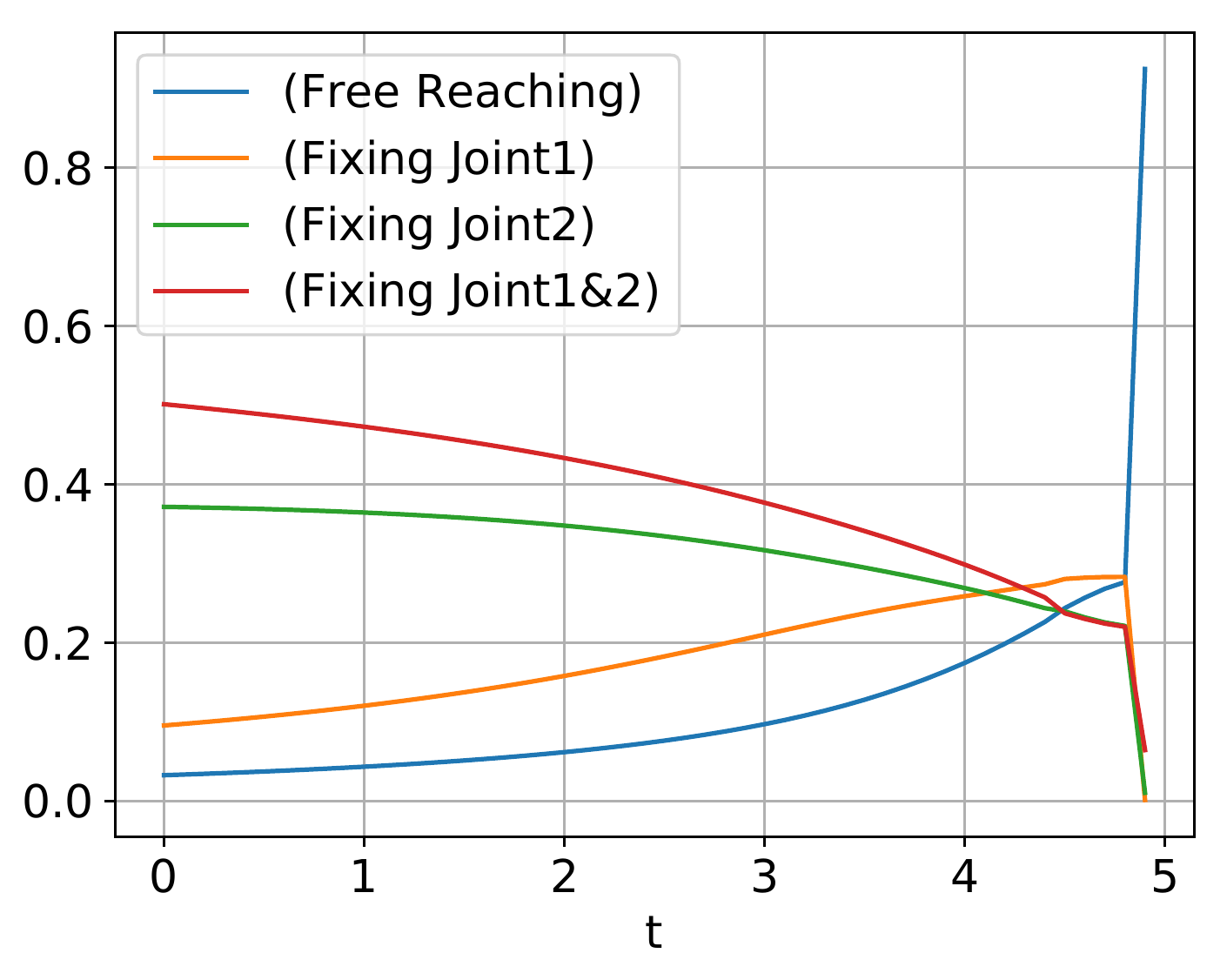}}	
	\subfigure[]{
		\includegraphics[width=.45\columnwidth, viewport = 5 10 400 315, clip]{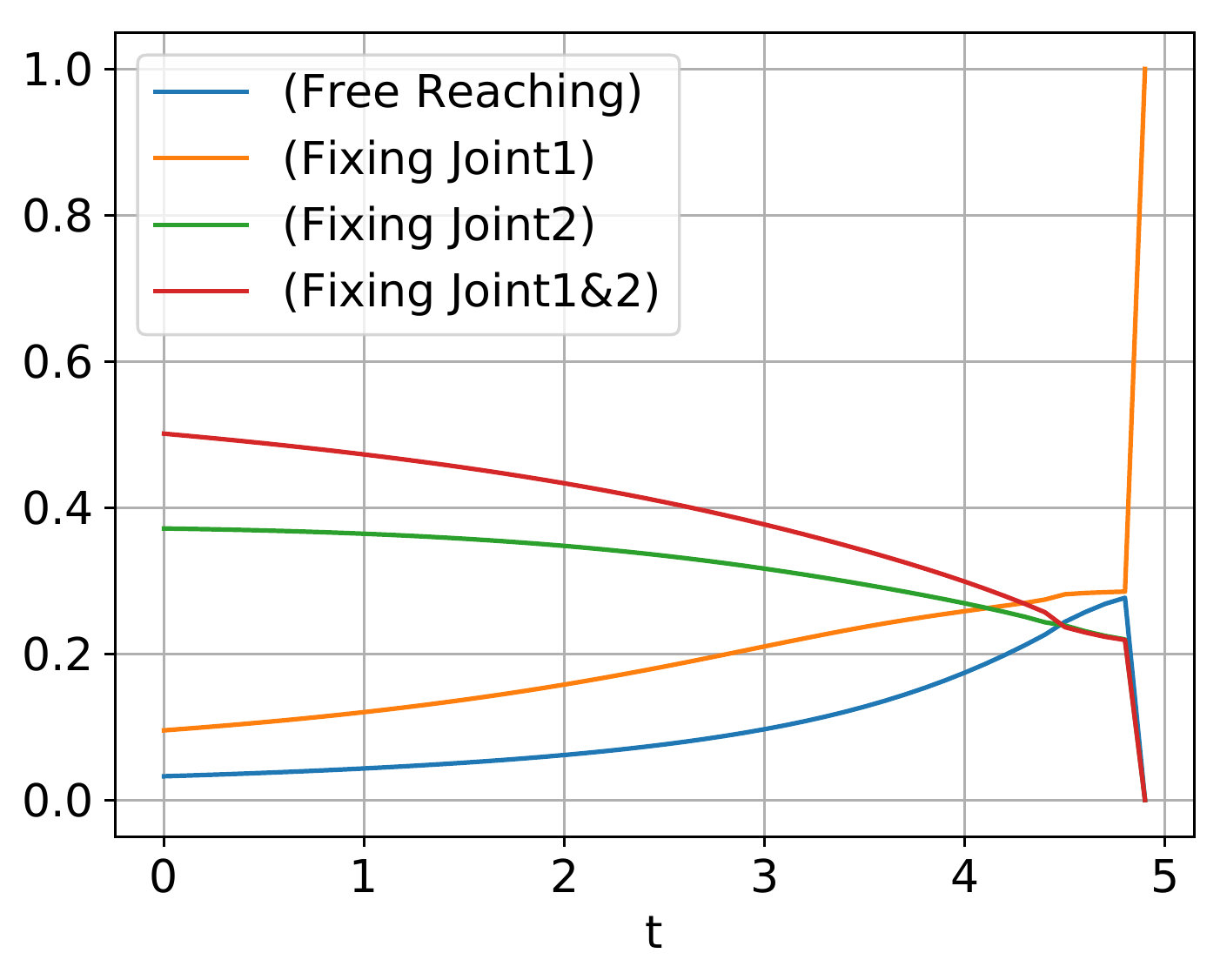}}	\vspace*{-.4cm}
	\caption{Two composite controllers: (a,c) blending, (b,d) switching.}\label{fig:control}\vspace*{-.5cm}
\end{figure}
\begin{figure*}[t]
	\centering
	\subfigure[initial config.]{
		\includegraphics[width=.25\columnwidth, viewport = 0 0 400 350, clip]{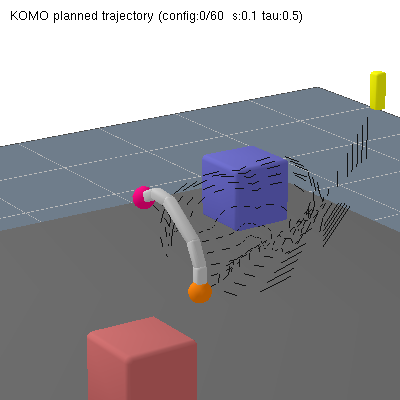}}	
	\subfigure[action 1: touch-endeffB-floor]{
		\includegraphics[width=.25\columnwidth, viewport = 0 0 400 350, clip]{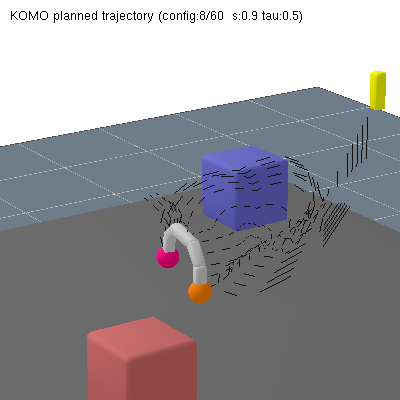}}	
	\subfigure[action 2: touch-endeffA-floor]{
		\includegraphics[width=.25\columnwidth, viewport = 0 0 400 350, clip]{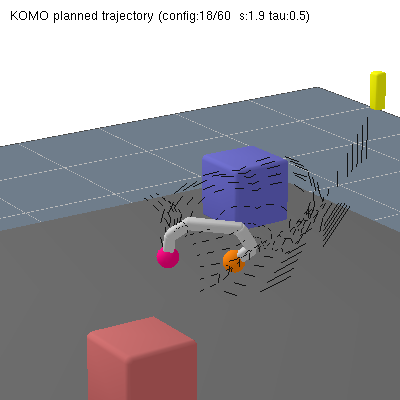}}	
	\subfigure[action 3: pick-endeffB-box1]{
		\includegraphics[width=.25\columnwidth, viewport = 0 0 400 350, clip]{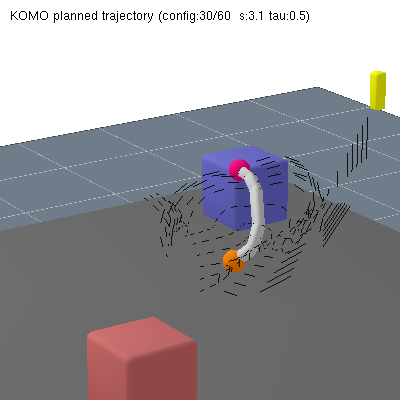}}	
	\subfigure[action 4: place-endeffB-box1]{
		\includegraphics[width=.25\columnwidth, viewport = 0 0 400 350, clip]{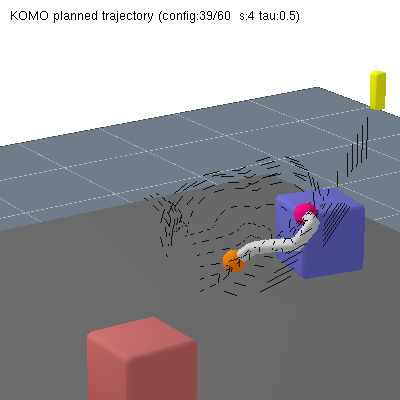}}	
	\subfigure[action 5: touch-endeffB-box1]{
		\includegraphics[width=.25\columnwidth, viewport = 0 0 400 350, clip]{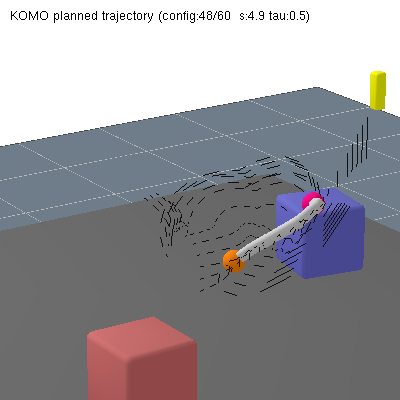}}	
	\subfigure[final config.]{
		\includegraphics[width=.25\columnwidth, viewport = 0 0 400 350, clip]{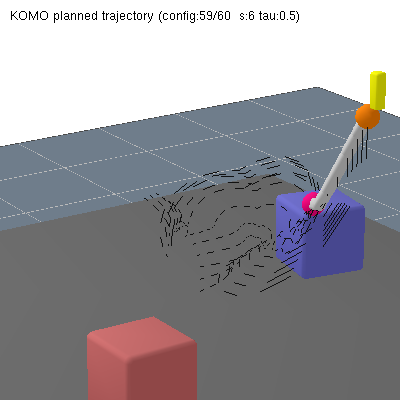}}	
	
	\subfigure[initial config.]{
		\includegraphics[width=.25\columnwidth, viewport = 0 0 400 350, clip]{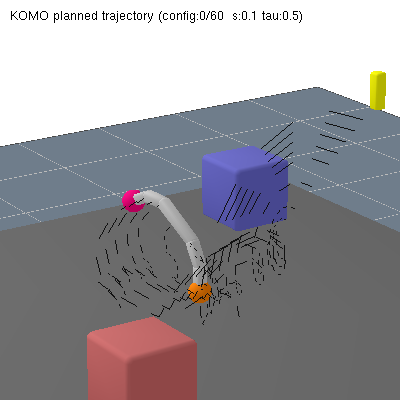}}	
	\subfigure[action 1: touch-endeffB-floor]{
		\includegraphics[width=.25\columnwidth, viewport = 0 0 400 350, clip]{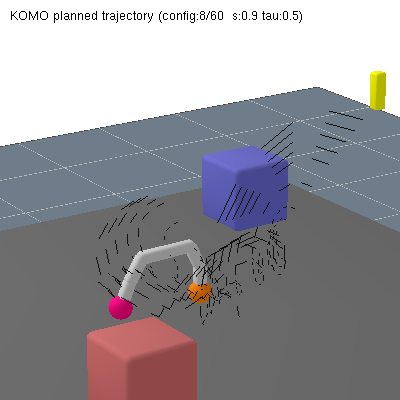}}	
	\subfigure[action 2: touch-endeffA-floor]{
		\includegraphics[width=.25\columnwidth, viewport = 0 0 400 350, clip]{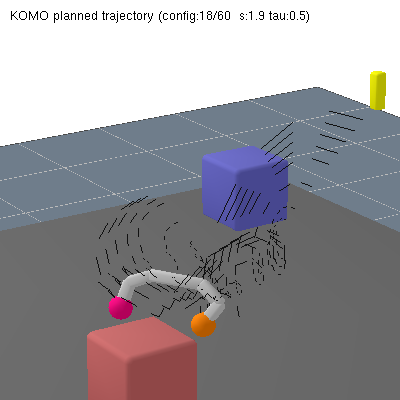}}	
	\subfigure[action 3: pick-endeffB-box2]{
		\includegraphics[width=.25\columnwidth, viewport = 0 0 400 350, clip]{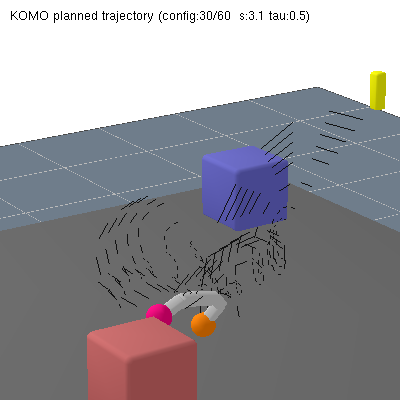}}	
	\subfigure[action 4: place-endeffB-box2]{
		\includegraphics[width=.25\columnwidth, viewport = 0 0 400 350, clip]{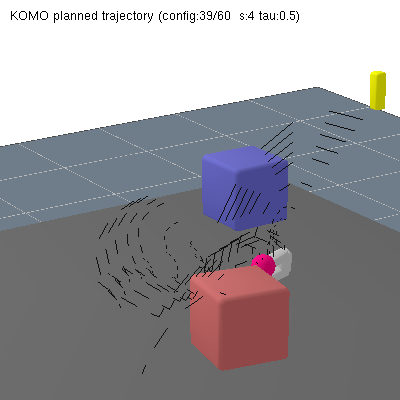}}	
	\subfigure[action 5: touch-endeffB-box2]{
		\includegraphics[width=.25\columnwidth, viewport = 0 0 400 350, clip]{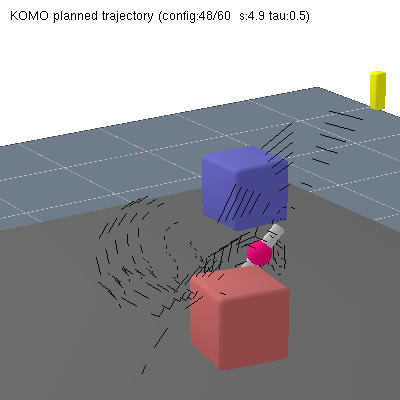}}	
	\subfigure[final config.]{
		\includegraphics[width=.25\columnwidth, viewport = 0 0 400 350, clip]{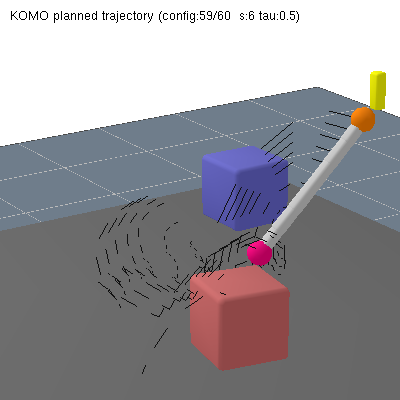}}\vspace*{-.2cm}	
	\caption{Two skeletons for the banana problem. Key frames for (a-g) \texttt{(Using Blue Box)} and for (h-i) \texttt{(Using Red Box)}.}\label{fig:banana}\vspace*{-.5cm}
\end{figure*}
\begin{figure}[t]
	\centering
	\includegraphics[width=.45\columnwidth, viewport = 5 10 400 315, clip]{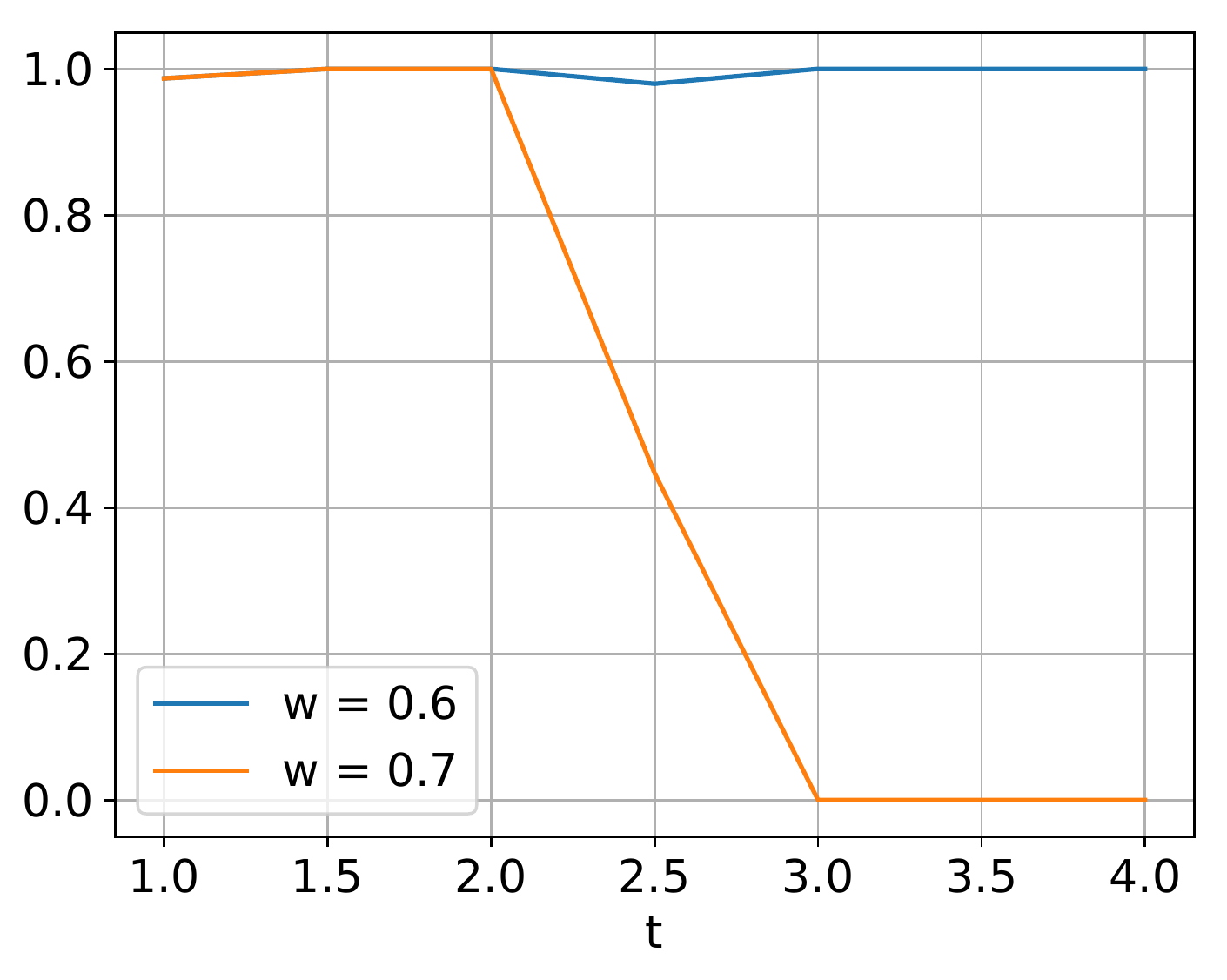}\vspace*{-.3cm}
	\caption{The mixture weight of \texttt{(Using Blue Box)} along time horizon when different disturbances were injected at $t=2$.}\label{fig:control2}\vspace*{-.3cm}
\end{figure}

\subsection{Contact exploitation: Elbow-on-table}
The first example considers a robot arm on a table having 4 degrees of freedom.
The dynamics of the robot is modeled as a double integrator, where control inputs and disturbances are directly injected as the acceleration of each joint. The goal of this task is to reach a target (red stars in Fig. \ref{fig:reach}) with its end-effector so the objective function penalizes the squared distance of the end-effector position to the target at the final time as well as the squared control cost along the time horizon ($T=5$). The problem also involves the inequality constraints to prevent the penetration of robot's body into the table. With the LGP formulation, four different skeletons are considered: reaching the target without touching the table, or while fixing one or two joints on the table for $t\in[3,5]$, i.e., the inequality constraints for the distance between those joints and the table are activated during that period. Fig. \ref{fig:reach} depicts the optimized trajectories with the sample paths from the optimal path distribution. We can observe that, as more degrees of freedom are restricted, the motions are more constrained but the uncertainties are substantially reduced. By constraining its configurations onto the constraint manifold, the robot becomes able to ``reject" some disturbances propagating to the end-effector.
This is quite realistic, given that tactile sensing from the contact makes it possible to maintain certain contact activities without having the high-gain feedback. Quantitative results are reported in Table~\ref{tab:Comp1}. As already discussed, \texttt{(Free Reaching)} has the lowest deterministic cost, while \texttt{(Fixing Joint1\&2)} minimizes the stochastic cost. By taking both costs into account as \eqref{eq:mixture_weight}, the robot can find the optimal trade-off and choose the best strategy, \texttt{(Fixing Joint2)}.

\begin{table}
	\renewcommand{\arraystretch}{1.3}
	\caption{Comparisons for reaching, $J^*\approx 2.9180$}\vspace*{-.4cm}
	\label{tab:Comp1}
	\centering
	\begin{tabular}{c||ccc}
		\hline
		&   $f(\mathbf{x}^*)$ 	  &$\left(|\Sigma^*_i|_+/|\Sigma_i|_+\right)^{1/2}$	& $p(\mathbf{a}^*)$  \\
		\hline\hline
		\scriptsize\texttt{(Free Reaching)}					&  \bf 0.1930 & 0.0041   &  0.0626 	       \\
		\scriptsize\texttt{(Fixing Joint1)}					&   0.7682 	  & 0.0099   	&  0.0850       \\
		\scriptsize\texttt{(Fixing Joint2)}  		 	   	& 0.6204     &  0.0584 		& \bf0.5810     	    \\
		\scriptsize\texttt{(Fixing Joint1\&2)} 		 	    &  1.4827     &  \bf0.0646   	& 0.2713	   \\
		\hline
	\end{tabular}\vspace*{-.6cm}
\end{table}

\subsection{Robust planning: Single- vs two-finger push}
The second example involves an object to be manipulated, where the goal is to push the object into the target position/orientation using one or two fingers. The robot's dynamics is modeled as a double integrator with 7 degrees of freedom and the motion of the pushed object is defined by the quasi-static dynamics~\cite{mason1986mechanics,zhou2018convex,toussaint20describing}. Under the optimal policy, both plans result in similar deterministic costs, and similarly small RMS errors of the final box configuration, $1.6568\times 10^{-5}$ and  $1.9720\times 10^{-5}$, respectively. Fig. \ref{fig:push_comp} shows that the two-finger push is inherently more stable, thereby having a lower stochastic cost; the robust strategy can be chosen only when the stochastic cost is also considered.

\subsection{Reactive controller: Elbow-on-table \& Banana} \label{sec:exp_reactive}
For the \textit{Elbow-on-table} example, Fig. \ref{fig:control} shows the executed trajectories with two composite control laws, blending \eqref{eq:blending} and switching \eqref{eq:switching}. We reduced the control cost weight for KODP to encourage the switching behavior. In both cases, the robot chooses the most constrained strategy, \texttt{(Fixing Joint1\&2)}, in earlier phases to reduce the uncertainty and shifts to less constrained modes. The blending controller, however, does not make the elbow completely put on the table while the switching does; the robot cannot fully exploit the uncertainty reduction benefit of \texttt{(Fixing Joint1\&2)} because of this undesirable smoothing effect.

The last example is a so-called banana problem; to catch a banana that is high up, a robot has to move a box first and then climb on it.
As depicted in Fig. \ref{fig:banana}, the robot in the considered scenario can use either the blue or the red box, and \texttt{(Using Blue Box)} has a lower cost since it is closer. We injected disturbances in the direction of the red box before the robot takes the first step, and considered the switching composition scheme. Fig. \ref{fig:control2} shows that, if the disturbance is small, the robot stays in \texttt{(Using Blue Box)}, but switches to \texttt{(Using Red Box)} if it is large.

\section{Conclusion}
This work has proposed a probabilistic framework for manipulation planning in stochastic domains. By connecting hybrid trajectory optimization and approximate posterior inference, we have built the optimal path distribution as a mixture of Gaussians. The proposed framework can evaluate plans not only in the deterministic sense but also in the sense of robustness, allowing for a reactive composite controller.

There is a close connection between this work and LQR-trees ~\cite{tedrake2010lqr}. By expanding a backward tree from the goal like LQR-trees, the reactive controller would become able to consider various plans more efficiently. Also, the exponential combinatorial complexity of skeletons (and thus the number of mixture components) can be addressed using deep architectures like~\cite{driess20deep} while the proposed method also can provide more sensible measures for learning such architectures.

\addtolength{\textheight}{-12cm}   




\section*{ACKNOWLEDGMENT}
We thank the MPI for Intelligent Systems for the Max Planck Fellowship funding.


\bibliographystyle{IEEEtran}
\bibliography{icra20_PLGP}

\begin{thebibliography}{10}
\providecommand{\url}[1]{#1}
\csname url@rmstyle\endcsname
\providecommand{\newblock}{\relax}
\providecommand{\bibinfo}[2]{#2}
\providecommand\BIBentrySTDinterwordspacing{\spaceskip=0pt\relax}
\providecommand\BIBentryALTinterwordstretchfactor{4}
\providecommand\BIBentryALTinterwordspacing{\spaceskip=\fontdimen2\font plus
\BIBentryALTinterwordstretchfactor\fontdimen3\font minus
  \fontdimen4\font\relax}
\providecommand\BIBforeignlanguage[2]{{%
\expandafter\ifx\csname l@#1\endcsname\relax
\typeout{** WARNING: IEEEtran.bst: No hyphenation pattern has been}%
\typeout{** loaded for the language `#1'. Using the pattern for}%
\typeout{** the default language instead.}%
\else
\language=\csname l@#1\endcsname
\fi
#2}}

\bibitem{lavalle1998rapidly}
S.~M. LaValle, ``Rapidly-exploring random trees: A new tool for path
  planning,'' 1998.

\bibitem{karaman2011sampling}
S.~Karaman and E.~Frazzoli, ``Sampling-based algorithms for optimal motion
  planning,'' \emph{The International Journal of Robotics Research}, vol.~30,
  no.~7, pp. 846--894, 2011.

\bibitem{mayne1966second}
D.~Mayne, ``A second-order gradient method for determining optimal trajectories
  of non-linear discrete-time systems,'' \emph{International Journal of
  Control}, vol.~3, no.~1, pp. 85--95, 1966.

\bibitem{todorov2005generalized}
E.~Todorov and W.~Li, ``A generalized iterative lqg method for locally-optimal
  feedback control of constrained nonlinear stochastic systems,'' in
  \emph{Proceedings of the 2005, American Control Conference, 2005.}\hskip 1em
  plus 0.5em minus 0.4em\relax IEEE, 2005, pp. 300--306.

\bibitem{toussaint2017tutorial}
M.~Toussaint, ``A tutorial on newton methods for constrained trajectory
  optimization and relations to slam, gaussian process smoothing, optimal
  control, and probabilistic inference,'' in \emph{Geometric and numerical
  foundations of movements}.\hskip 1em plus 0.5em minus 0.4em\relax Springer,
  2017, pp. 361--392.

\bibitem{mordatch2012discovery}
I.~Mordatch, E.~Todorov, and Z.~Popovi{\'c}, ``Discovery of complex behaviors
  through contact-invariant optimization,'' \emph{ACM Transactions on Graphics
  (TOG)}, vol.~31, no.~4, p.~43, 2012.

\bibitem{deits2014footstep}
R.~Deits and R.~Tedrake, ``Footstep planning on uneven terrain with
  mixed-integer convex optimization,'' in \emph{2014 IEEE-RAS International
  Conference on Humanoid Robots}.\hskip 1em plus 0.5em minus 0.4em\relax IEEE,
  2014, pp. 279--286.

\bibitem{toussaint2018differentiable}
M.~Toussaint, K.~Allen, K.~A. Smith, and J.~B. Tenenbaum, ``Differentiable
  physics and stable modes for tool-use and manipulation planning.'' in
  \emph{Robotics: Science and Systems}, 2018.

\bibitem{todorov2008general}
E.~Todorov, ``General duality between optimal control and estimation,'' in
  \emph{2008 47th IEEE Conference on Decision and Control}.\hskip 1em plus
  0.5em minus 0.4em\relax IEEE, 2008, pp. 4286--4292.

\bibitem{rawlik2012stochastic}
K.~Rawlik, M.~Toussaint, and S.~Vijayakumar, ``On stochastic optimal control
  and reinforcement learning by approximate inference,'' in \emph{Robotics:
  Science and Systems}, 2012, p. 3052–3056.

\bibitem{kappen2012optimal}
H.~J. Kappen, V.~G{\'o}mez, and M.~Opper, ``Optimal control as a graphical
  model inference problem,'' \emph{Machine learning}, vol.~87, no.~2, pp.
  159--182, 2012.

\bibitem{toussaint2009robot}
M.~Toussaint, ``Robot trajectory optimization using approximate inference,'' in
  \emph{Proceedings of the 26th annual international conference on machine
  learning}.\hskip 1em plus 0.5em minus 0.4em\relax ACM, 2009, pp. 1049--1056.

\bibitem{todorov2009efficient}
E.~Todorov, ``Efficient computation of optimal actions,'' \emph{Proceedings of
  the national academy of sciences}, vol. 106, no.~28, pp. 11\,478--11\,483,
  2009.

\bibitem{kappen2016adaptive}
H.~J. Kappen and H.~C. Ruiz, ``Adaptive importance sampling for control and
  inference,'' \emph{Journal of Statistical Physics}, vol. 162, no.~5, pp.
  1244--1266, 2016.

\bibitem{ha2018adaptive}
J.-S. Ha, Y.-J. Park, H.-J. Chae, S.-S. Park, and H.-L. Choi, ``Adaptive
  path-integral autoencoders: Representation learning and planning for
  dynamical systems,'' in \emph{Advances in Neural Information Processing
  Systems}, 2018, pp. 8927--8938.

\bibitem{gardiner1985handbook}
C.~W. Gardiner \emph{et~al.}, \emph{Handbook of stochastic methods}.\hskip 1em
  plus 0.5em minus 0.4em\relax Springer Berlin, 1985, vol.~4.

\bibitem{ha2016topology}
J.-S. Ha and H.-L. Choi, ``A topology-guided path integral approach for
  stochastic optimal control,'' in \emph{2016 IEEE International Conference on
  Robotics and Automation (ICRA)}.\hskip 1em plus 0.5em minus 0.4em\relax IEEE,
  2016, pp. 4605--4612.

\bibitem{erez2012trajectory}
T.~Erez and E.~Todorov, ``Trajectory optimization for domains with contacts
  using inverse dynamics,'' in \emph{2012 IEEE/RSJ International Conference on
  Intelligent Robots and Systems}.\hskip 1em plus 0.5em minus 0.4em\relax IEEE,
  2012, pp. 4914--4919.

\bibitem{zucker2013chomp}
M.~Zucker, N.~Ratliff, A.~D. Dragan, M.~Pivtoraiko, M.~Klingensmith, C.~M.
  Dellin, J.~A. Bagnell, and S.~S. Srinivasa, ``Chomp: Covariant hamiltonian
  optimization for motion planning,'' \emph{The International Journal of
  Robotics Research}, vol.~32, no. 9-10, pp. 1164--1193, 2013.

\bibitem{toussaint2017multi}
M.~Toussaint and M.~Lopes, ``Multi-bound tree search for logic-geometric
  programming in cooperative manipulation domains,'' in \emph{2017 IEEE
  International Conference on Robotics and Automation (ICRA)}.\hskip 1em plus
  0.5em minus 0.4em\relax IEEE, 2017, pp. 4044--4051.

\bibitem{gelman2013bayesian}
A.~Gelman, J.~B. Carlin, H.~S. Stern, D.~B. Dunson, A.~Vehtari, and D.~B.
  Rubin, \emph{Bayesian data analysis}.\hskip 1em plus 0.5em minus 0.4em\relax
  Chapman and Hall/CRC, 2013.

\bibitem{levy1995sensitivity}
A.~Levy and R.~Rockafellar, ``Sensitivity of solutions in nonlinear programming
  problems with nonunique multipliers,'' \emph{Recent Advances in Nonsmooth
  Optimization}, p. 215, 1995.

\bibitem{amos2017optnet}
B.~Amos and J.~Z. Kolter, ``Optnet: Differentiable optimization as a layer in
  neural networks,'' in \emph{Proceedings of the 34th International Conference
  on Machine Learning-Volume 70}.\hskip 1em plus 0.5em minus 0.4em\relax JMLR.
  org, 2017, pp. 136--145.

\bibitem{todorov2009compositionality}
E.~Todorov, ``Compositionality of optimal control laws,'' in \emph{Advances in
  Neural Information Processing Systems}, 2009, pp. 1856--1864.

\bibitem{muico2011composite}
U.~Muico, J.~Popovi{\'c}, and Z.~Popovi{\'c}, ``Composite control of physically
  simulated characters,'' \emph{ACM Transactions on Graphics (TOG)}, vol.~30,
  no.~3, pp. 1--11, 2011.

\bibitem{mason1986mechanics}
M.~T. Mason, ``Mechanics and planning of manipulator pushing operations,''
  \emph{The International Journal of Robotics Research}, vol.~5, no.~3, pp.
  53--71, 1986.

\bibitem{zhou2018convex}
J.~Zhou, M.~T. Mason, R.~Paolini, and D.~Bagnell, ``A convex polynomial model
  for planar sliding mechanics: theory, application, and experimental
  validation,'' \emph{The International Journal of Robotics Research}, vol.~37,
  no. 2-3, pp. 249--265, 2018.

\bibitem{toussaint20describing}
M.~Toussaint, J.-S. Ha, and D.~Driess, ``Describing physics for physical
  reasoning: Force-based sequential manipulation planning,'' \emph{arXiv
  preprint arXiv:2002.12780}, 2020.

\bibitem{tedrake2010lqr}
R.~Tedrake, I.~R. Manchester, M.~Tobenkin, and J.~W. Roberts, ``{LQR}-trees:
  Feedback motion planning via sums-of-squares verification,'' \emph{The
  International Journal of Robotics Research}, vol.~29, no.~8, pp. 1038--1052,
  2010.

\bibitem{driess20deep}
D.~Driess, O.~Oguz, J.-S. Ha, and M.~Toussaint, ``Deep visual heuristics:
  Learning feasibility of mixed-integer programs for manipulation planning,''
  in \emph{{IEEE} International Conference on Robotics and Automation (ICRA)},
  2020.

\end{thebibliography}

\end{document}